\documentclass[journal]{IEEEtran}

\usepackage{cite}
\usepackage{amsmath,amssymb}
\usepackage{graphicx}
\usepackage{booktabs}
\usepackage{multirow}
\usepackage{array}
\usepackage{tabularx}
\usepackage{makecell}
\usepackage{url}
\usepackage{balance}
\usepackage{float}
\usepackage{placeins}
\usepackage{nccmath} 
\usepackage[
  colorlinks=true,
  linkcolor=blue,
  citecolor=blue,
  urlcolor=blue
]{hyperref}

\usepackage{orcidlink}

\graphicspath{{figures/}}
\begin{document}

\title{COBICount: Separating Object and Background Responses for Remote Sensing Object Counting Without Training on Target Data}

\newcommand{\orcidauthorA}{0009-0002-4631-1018} 
\newcommand{\orcidauthorB}{0009-0007-8171-1478} 
\newcommand{\orcidauthorC}{0000-0002-8836-1382} 
\newcommand{\orcidauthorD}{0009-0004-5020-8464} 

\author{%
Junjing Zheng\,\orcidlink{\orcidauthorB},
Zhiyi Zhou\,\orcidlink{\orcidauthorA},
Ningrui Yang\,\orcidlink{\orcidauthorD},
and Hongying Meng\,\orcidlink{\orcidauthorC}, \textit{Senior Member, IEEE}
\thanks{Junjing Zheng is with the School of International Studies, Chongqing University of Posts and Telecommunications, Chongqing 400065, China, and also with Brunel University London, Uxbridge UB8 3PH, U.K. (email: 2378933@brunel.ac.uk).}%
\thanks{Zhiyi Zhou is with the School of International Studies, Chongqing University of Posts and Telecommunications, Chongqing 400065, China, and also with Brunel University London, Uxbridge UB8 3PH, U.K. (email: 2378903@brunel.ac.uk).}%
\thanks{Ningrui Yang is with the School of International Studies, Chongqing University of Posts and Telecommunications, Chongqing 400065, China, and also with Brunel University London, Uxbridge UB8 3PH, U.K. (email: 2378911@brunel.ac.uk).}%
\thanks{Hongying Meng is with the Department of Electronic and Electrical Engineering, College of Engineering, Design and Physical Sciences, Brunel University of London, Uxbridge UB8 3PH, U.K. (email: hongying.meng@brunel.ac.uk).}%
\thanks{\textit{(Corresponding authors: Junjing Zheng; Hongying Meng.)}}
\thanks{Preprint. This manuscript has been submitted to IEEE JSTARS
for possible publication.}
}

\markboth{Preprint, Zheng et al.: COBICount}%
{Zheng: COBICount: Remote Sensing Object Counting without Training on Target Data}

\maketitle


\begin{abstract}
Remote sensing object counting estimates how many buildings, vehicles, or ships appear in overhead images. Most supervised counters predict a density map, whose sum gives the object count, and assume similar categories, sizes, and backgrounds. Applying them across regions, sensors, or categories often requires target data or further training, which may be costly or unavailable. We study source-only counting. Training for the counting task and model selection use one group of images that shares an object category and similar imaging conditions, with one point marking each object. Target images and information remain unavailable until the model is fixed. This reduces data preparation but makes transfer harder. A model trained on one source may place high density values, called responses, on real objects and repeated background structures. Road edges, parking grids, roof boundaries, and water boundaries may then be counted as objects, creating candidate origin ambiguity. COBICount separates response generation, acceptance, and background suppression. Candidate Evidence (CE) generates possible responses. Candidate Acceptance (CA) keeps compact responses centered on objects. Bias Isolation (BI) reduces responses associated with repeated background structures. Their outputs form the final density map. Trained on RSOC Building and evaluated directly on DOTA Large Vehicle, Small Vehicle, and Ship, COBICount achieves the lowest mean absolute error (MAE) averaged over the target domains among the compared methods, 174.132. It uses 5.07 million parameters and 17.41 billion floating point operations for a $512\times512$ input. COBICount improves transfer without target data or training for each target. The code will be available at: \url{https://github.com/yixuxi22/COBICount}.
\end{abstract}


\section{Introduction}
\label{sec:introduction}

\IEEEPARstart{R}{emote} sensing object counting estimates how many instances of a selected class, such as buildings, vehicles, or ships, appear in an aerial or satellite image. It supports urban mapping, traffic monitoring, maritime surveillance, and infrastructure assessment. One approach detects each object and then counts the detections~\cite{xia2018dota,lam2018xview}. Another uses point supervision, where each training object is marked by one point. A counter learns a nonnegative density map that distributes the object count across spatial locations; summing all map values gives the number of objects in the image~\cite{zhang2016mcnn,li2018csrnet,gao2020counting}. Point annotations require less detail than bounding boxes, which makes them useful for scenes containing many objects. Remote sensing images remain difficult because their overhead views, image resolutions, object sizes, and background layouts vary widely. Accurate counting requires both a correct total and high local density values placed on true objects. We refer to these local values as counting responses.

Most supervised counters are developed in a setting where training and test images come from the same domain and contain similar object classes, sizes, and scene patterns. Methods based on density estimation, point prediction, local context, and transformers have improved counting under this condition~\cite{liu2019can,ma2019bayesian,wang2020dmcount,song2021p2pnet,liang2022transcrowd}. Remote sensing methods also address changes in object size and complex backgrounds within annotated datasets~\cite{gao2022psgcnet,guo2022tasnet,shen2025edgecount}. This setting is useful for a fixed task, but it can hide dependence on the training data. A model trained on buildings may learn not only building appearance but also roof edges, block layouts, and nearby roads. These patterns may change when the model is applied to another region, sensor, image resolution, or object class. Good accuracy on the annotated training data does not show whether the same responses will remain valid after such a change.

When training and deployment images differ in object category or imaging conditions, they belong to different \emph{domains}. A domain is a group of images that share the same object category and similar imaging and scene conditions; it is not an ordinary training, validation, or test partition. The \emph{source domain} provides the point annotations used during model development. Its training split updates the model, its validation split selects the final model, and a separate test split with the same category and imaging conditions still measures performance in the source domain. A \emph{target domain} has a different object category or different imaging conditions and is unavailable during model development.

Existing approaches to transfer between domains use different forms of additional data. Domain adaptation requires labeled or unlabeled target images during training~\cite{tuia2016domain,ganin2016dann}. Domain generalization does not use target images during training, but its conventional form often requires several annotated source domains~\cite{li2018mldg,mansilla2021domain}. Pretrained vision models first learn from large external datasets and can later provide reusable features or masks~\cite{radford2021clip,kirillov2023sam,liu2024remoteclip,zhang2024georsclip}. These outputs do not directly form a density map learned from point annotations or show whether a local counting response lies on a true object. These data requirements can prevent deployment when representative target images cannot be collected in advance, several annotated source domains are unavailable, or adaptation for every new region, sensor, or object category is impractical.

For this reason, training for the counting task and model selection use only one source domain, with one point annotation for each object and no target data. This setting reduces the data that must be prepared before deployment and allows a fixed model to be applied to new domains without further training. The choice of source and target domains depends on the experimental protocol and is not fixed. In the main experiment, RSOC Building is used as the source domain, and DOTA Large Vehicle, Small Vehicle, and Ship are used as three unseen target domains. Other experiments use different combinations of source and target domains; for example, RSOC Building remains the source domain when DIOR Airplane is evaluated as an unseen target domain. We call this setting \emph{source-only counting}. It limits the data used for the counting task. A generic initialization fixed before source training may be used when it is part of a baseline's standard implementation, but it is neither selected nor updated with target data. No target images or information derived from them, including labels, automatically generated labels, feature summaries, values used to normalize the data, validation results, or signals used to choose the model, are used before the final source model is fixed. Its parameters are not changed afterward. Target images are then used only to produce predictions, and target annotations are used only by the evaluator to calculate performance metrics.

This reduced dependence on target data makes transfer more difficult. Source point annotations mark object centers, but they do not label roads, parking spaces, roof boundaries, water boundaries, or other background regions. The model may respond to both the annotated objects and background structures that repeatedly occur in the source images. After the domain changes, one high response may be correct because it is centered on a real target object and forms the compact local pattern encouraged around source point annotations. Another high response may be incorrect because a road edge, parking grid, roof boundary, or harbor structure forms a similar pattern. The response value alone does not reveal which case produced it. We call this problem \emph{candidate origin ambiguity}. A change in the spatial size of a response adds another difficulty. A source building may cover a broad area in the network output, whereas a small target vehicle may occupy only a few locations. The vehicle response can then be weaker than nearby parking or road patterns. Mean absolute error (MAE) and root mean squared error (RMSE) cannot fully reveal these failures because they measure only the total count. Missed objects and false background responses may partly cancel.

We propose \textbf{COBICount} to decide which local responses should contribute to the count. Rather than asking one prediction branch to generate and accept every response at once, COBICount separates response generation, acceptance, and background suppression. Candidate Evidence (CE) produces a broad nonnegative map of possible responses so that weak object responses can enter the counting process. Candidate Acceptance (CA) keeps responses that match the compact and centered patterns learned around source point annotations. Bias Isolation (BI) reduces responses associated with learned background structures, edges, broad responses not centered on objects, and repeated grids. A fixed mask identifies valid image regions and removes the black padding outside them. These outputs form the final density map, whose sum gives the predicted count. All training targets are derived from source point annotations. Regions near these annotations show where responses should remain, while valid image regions farther away show where responses should be suppressed. BI does not identify named background categories in a target domain. An auxiliary inspection head, called Audit, is trained only with source annotations and is used to inspect response patterns after training. It does not control the final density map or use target annotations.

The main contributions are summarized as follows:
\begin{itemize}
    \item We study remote sensing object counting under \emph{source-only counting}. We clarify that training, validation, and test subsets with the same object category and imaging conditions belong to the same domain. Under this setting, candidate origin ambiguity explains why a model trained on one annotated source domain may produce incorrect counting responses in an unseen target domain.

    \item We introduce COBICount, which separates candidate generation, local acceptance, and background suppression through CE, CA, and BI. All supervision is derived from source point annotations. The model requires neither adaptation using target data nor labels for target background categories.

    \item We evaluate count accuracy and response location through a main comparison that includes recent generic baselines and baselines for remote sensing evaluated under source-only counting, tests that change the source domain, ablations, checks of whether predictions are too high or too low and whether responses in the final density map occur near reference points. When RSOC Building is the source and the three DOTA categories are unseen targets, COBICount obtains the lowest MAE averaged over the target domains among the compared methods.
\end{itemize}

\section{Related Work}
\label{sec:related_work}

The studies most closely related to COBICount address three questions. First, how can remote sensing objects be counted when each training object is marked by only one point? Second, what data do existing methods require when training and deployment domains differ? Third, how can we check whether a counting response is located on an object rather than on a background structure? The following review discusses these questions in this order.

\subsection{Counting with Point Supervision in Remote Sensing Images}
\label{subsec:rw_rs_counting_density}

Remote sensing objects can be counted through detection or point supervision. Detection methods first locate individual objects and then use the number of detections as the count. DOTA~\cite{xia2018dota}, xView~\cite{lam2018xview}, DIOR~\cite{li2020dior}, and FAIR1M~\cite{sun2022fair1m} provide bounding box annotations for objects in overhead images. CARPK and PUCPR+ focus on vehicles observed from elevated viewpoints~\cite{hsieh2017carpk}. Bounding boxes describe both object position and extent, but collecting them is costly when an image contains many small objects. Counting with detection can also inherit errors from missed and duplicate detections.

Point supervision reduces this annotation burden by marking only one center point for each object. A common approach converts these points into a density map used as the training reference. A network predicts this map, and the count is obtained by summing its spatial values. Early crowd counting models established several parts of this formulation. The Multi-Column Convolutional Neural Network (MCNN) uses parallel convolutional columns to handle changes in object size~\cite{zhang2016mcnn}. CSRNet uses dilated convolutions to observe a larger image area without further reducing map resolution~\cite{li2018csrnet}. The Context Aware Network (CAN) selects information from the surrounding image for different regions~\cite{liu2019can}. Bayesian Loss accounts for uncertainty around point annotations~\cite{ma2019bayesian}, and a distribution matching method compares the spatial distributions of predicted and reference density values~\cite{wang2020dmcount}. Other formulations predict object points directly, as in P2PNet~\cite{song2021p2pnet}, or use transformers to relate distant image regions, as in TransCrowd~\cite{liang2022transcrowd}.

Objects in remote sensing images can vary greatly in size, and their backgrounds often contain repeated roads, roofs, parking lines, and boundaries. The Remote Sensing Object Counting (RSOC) dataset introduced subsets for buildings, ships, large vehicles, and small vehicles, with one point marking each object~\cite{gao2020counting}. The Pyramidal Scale and Global Context Guided Network (PSGCNet) combines features computed at several image resolutions with information from the complete scene~\cite{gao2022psgcnet}. The Triple Attention and Scale Aware Network (TASNet) uses separate processing for different object sizes and directs the model toward useful features~\cite{guo2022tasnet}. The Balanced Density Regression Network (BDRNet) combines density regression with an auxiliary object region output that predicts the image regions occupied by objects~\cite{guo2024bdrnet}. EdgeCount transfers density map knowledge from a larger network to a smaller network to reduce computation while retaining counting accuracy~\cite{shen2025edgecount}. Other remote sensing tasks report related image difficulties. A study that estimates surface depth from two satellite views addresses changes in image resolution~\cite{wang2025mscanet}, and a study that labels water pixels in radar and optical images covers broad geographic areas~\cite{wieland2024s1s2water}. These studies do not perform counting, but they show that object size, sensing method, and boundary appearance can change across overhead images.

Most counting methods above are trained and tested separately on each dataset or object category. They explain how to produce an accurate count when training and test conditions are similar, but not how to decide whether a response remains valid after the category, sensor, resolution, or background changes. In particular, a low count error does not show whether high values in the predicted density map fall on the objects or on repeated background structures in the source images.

\subsection{Learning across Domains under Different Data Conditions}
\label{subsec:rw_dg_foundation}

Methods for transfer between domains differ mainly in the data available before evaluation. Domain adaptation uses images from the target domain during model training, either with or without target labels. Adversarial domain training, for example, learns features that make source and target samples harder to distinguish~\cite{ganin2016dann}. Remote sensing studies have used adaptation to address changes in sensors, regions, and imaging conditions~\cite{tuia2016domain}, including unsupervised adaptation for image segmentation, which assigns a class to each pixel~\cite{ma2024deglgan}. Another adaptation setting removes access to the original source images during adaptation but still uses target images; remote sensing object detection provides one example~\cite{liu2024sfodrs}. Domain adaptation can reduce a known difference between source and target domains, but it requires target samples before deployment and usually repeats adaptation for each new target domain.

Domain generalization removes target domain images from training. Its conventional form instead learns from several annotated source domains. Meta Learning for Domain Generalization (MLDG) simulates domain changes by treating some available domains as temporary training domains and others as temporary test domains~\cite{li2018mldg}. A method called gradient surgery reduces conflicts among updates from different source domains~\cite{mansilla2021domain}. These methods avoid collecting target data, but several labeled source domains may also be difficult to obtain. Counting research has also considered more restricted training data. A domain general crowd counting method divides source crowd data into groups and separates information shared by the groups from information tied to each group~\cite{du2023domain}. MPCount studies crowd counting from one labeled domain and encourages the model to retain similar responses when source images are transformed~\cite{peng2024mpcount}. This use of one labeled source is related to source-only counting, although MPCount was developed for people counting across crowd datasets. Its evaluation under the present remote sensing protocol is described in Section~\ref{subsec:exp_metrics_comparison}. Universal Representation Matching (URM) uses features from models trained to connect images and language, and it counts categories specified by a small set of target examples~\cite{chen2025urm}. These methods improve generalization under limited counting data, but their original settings differ from ours. Crowd counting methods continue to count people, whereas methods that receive a small set of target examples use those examples to specify the category. Our setting fixes a model using one remote sensing source domain with one point per object and provides no target image or example before evaluation on new categories and imaging conditions.

Pretrained models offer another way to reduce the annotations required for a particular task. A pretrained model first learns from a large external collection and is then reused for a specific task. One model learns by matching image features with text descriptions~\cite{radford2021clip}, and the Segment Anything Model (SAM) produces image masks from prompts such as points or boxes~\cite{kirillov2023sam}. Pretraining for remote sensing also considers properties of overhead data. SatMAE hides parts of satellite images and learns to reconstruct them; it uses images captured at several times and wavelengths~\cite{cong2022satmae}. Another pretraining method explicitly models changes in geographic scale~\cite{reed2023scalemae}. SatMAE++ learns features at several spatial levels from images captured at multiple wavelengths~\cite{noman2024satmaepp}. CrossEarth studies features used for remote sensing segmentation across domains~\cite{gong2024crossearth}.

Models that connect vision and language have also been adapted to remote sensing tasks. RemoteCLIP and GeoRSCLIP learn links between overhead images and text~\cite{liu2024remoteclip,zhang2024georsclip}. A language guided detector uses text to specify categories rather than relying on a fixed training list~\cite{pan2024lae}. One study uses SAM to expand remote sensing segmentation data~\cite{wang2023samrs}, and RSPrompter learns prompts that help separate individual objects in remote sensing images~\cite{chen2023rsprompter}. Other studies test how SAM transfers with no remote sensing example or only one labeled example~\cite{osco2023samrsapp}. These models can supply features that describe image content, detections, or masks, but those outputs do not by themselves define a count. A feature may indicate that an image contains vehicles without stating how many vehicles are present. A mask may cover one object, several touching objects, or a background region. Pretrained models therefore do not remove the need to place an appropriate counting response on each object and reject responses caused by background structures.

\subsection{Checking Where Counting Responses Occur}
\label{subsec:rw_diagnosis_position}

MAE and RMSE compare the predicted and reference totals, but they do not show where the error occurs. A missed object and an extra response on the background can offset each other and leave a seemingly accurate total. This failure is related to \emph{shortcut learning}, in which a model reduces its training loss by using an easy recurring pattern rather than the intended object evidence~\cite{geirhos2020shortcut}. In remote sensing images, such patterns may include roof edges around buildings, road or parking layouts around vehicles, and harbor or water boundaries around ships. When only one source domain is annotated, the model has no direct label stating that these background structures should not be counted.

General explanation methods provide only part of the required evidence. A gradient method for class activation mapping highlights image regions that influence a network decision~\cite{selvaraju2017gradcam}, but its broad visual maps do not test whether a density map places a separate response on each object. An analysis designed for counting can instead compare response positions with annotated points. Responses near annotated points show where objects are covered, whereas responses far from every annotation reveal possible background responses. The fractions of annotated objects covered and responses supported by annotations can expose missed objects and false responses that total count errors may hide.

Existing studies mainly improve the predicted count or explain a trained model after the fact. COBICount connects model design with checking response positions. CE generates possible object responses, CA keeps responses that match compact and centered patterns learned from source point annotations, and BI reduces responses associated with repeated background structures. All three modules are trained from source images and point annotations. A later analysis checks whether the final responses occur near target objects, without changing the predicted count or supplying target domain information to the model.

\section{Method}
\label{sec:method}

\subsection{Problem Setting and Model Output}
\label{subsec:method_problem}

Source-only counting, introduced in Section~\ref{sec:introduction}, is written formally here. Let
\begin{equation}
    \mathcal{D}_{s}
    =
    \left\{
    \left(\mathbf I_i^{s},\mathcal P_i^{s}\right)
    \right\}_{i=1}^{N_s}
    \label{eq:source_dataset}
\end{equation}
be the source domain. It contains $N_s$ images. For image $\mathbf I_i^{s}$,
$\mathcal P_i^{s}=\{\mathbf p_{ij}\}_{j=1}^{n_i}$ contains one point for each of its $n_i$ objects. A point is written as $\mathbf p_{ij}=(x_{ij},y_{ij})$, where $x_{ij}$ is the horizontal coordinate and $y_{ij}$ is the vertical coordinate. When the point is used to access a tensor, the vertical coordinate gives the row index and the horizontal coordinate gives the column index. Therefore, $\mathbf X(\mathbf p_{ij})$ denotes the tensor entry $\mathbf X(y_{ij},x_{ij})$.

Training and model selection use only $\mathcal{D}_s$. Let $\mathcal{D}_t$ denote an unseen target domain as defined in Section~\ref{sec:introduction}. Images from $\mathcal{D}_t$ are used only after the model has been fixed. No target image, label, automatically generated label, feature summary, normalization value, validation result, or model selection signal is used to update or select the model.

Let $\mathbf I\in\mathbb R^{3\times H\times W}$ be an input image, where $H$ and $W$ are its height and width. Let $\Omega_{\mathbf I}$ denote the set of all pixel locations in $\mathbf I$:
\begin{equation}
    \Omega_{\mathbf I}
    =
    \{0,\ldots,W-1\}
    \times
    \{0,\ldots,H-1\}.
    \label{eq:image_grid_definition}
\end{equation}
Following the density map formulation described in Section~\ref{subsec:rw_rs_counting_density}, let $\widehat{\mathbf D}_{\mathbf I}$ be the predicted density map for $\mathbf I$. Its predicted count is
\begin{equation}
    \widehat N_{\mathbf I}
    =
    \sum_{\mathbf x\in\Omega_{\mathbf I}}
    \widehat{\mathbf D}_{\mathbf I}(\mathbf x).
    \label{eq:density_sum_count}
\end{equation}
For the single image considered below, we write $\widehat N$ for $\widehat N_{\mathbf I}$. The value at one location is not a class probability. It is only that location's contribution to the count. COBICount follows this summation rule, but it forms the density map through three consecutive decisions: generate possible responses, retain responses that agree with source point patterns, and reduce responses related to repeated structures.

\subsection{Overall Data Flow}
\label{subsec:method_overall_arch}

\begin{figure*}[!t]
    \centering
    \includegraphics[width=0.98\textwidth]{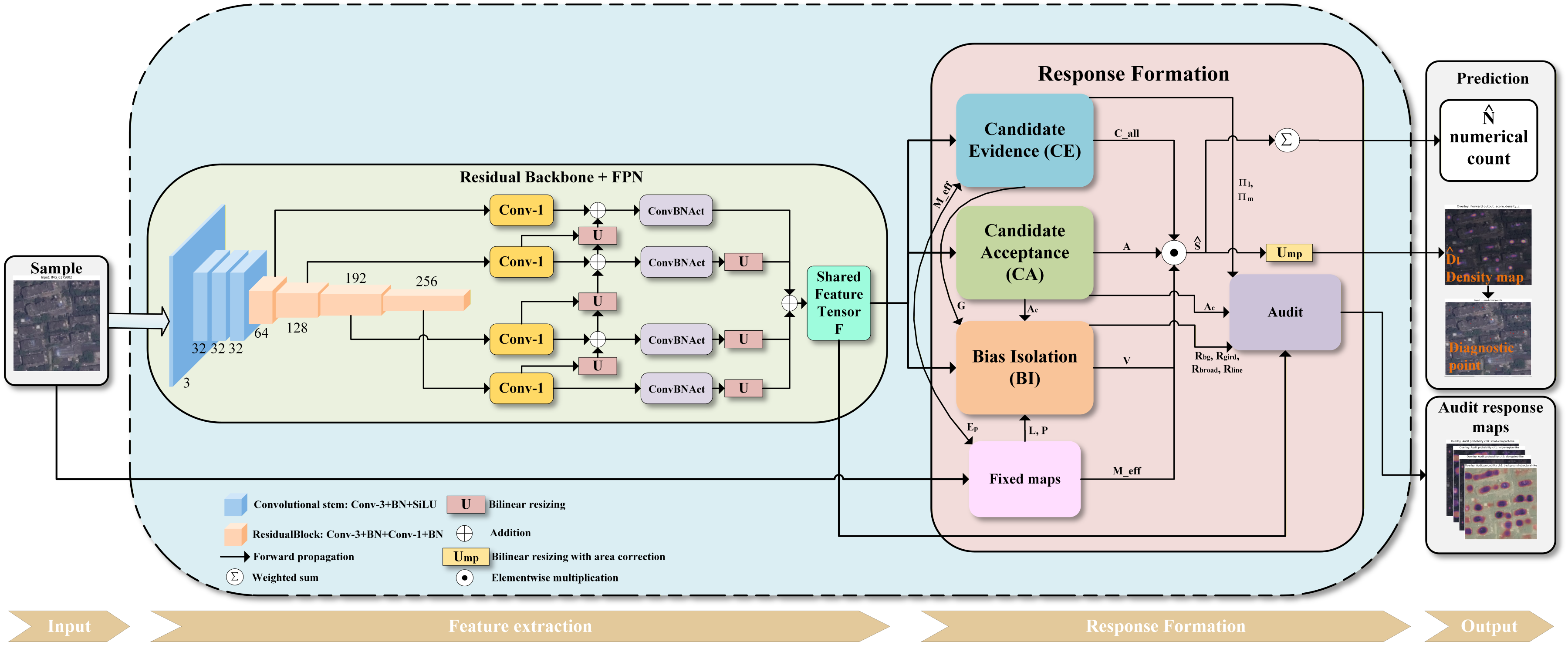}
    \caption{Overall data flow of COBICount. The input image produces the shared feature tensor $\mathbf F$ and the effective valid mask $\mathbf M^{eff}$. CE, CA, and BI produce $\mathbf C^{all}$, $\mathbf A$, and $\mathbf V$. These four maps enter one elementwise multiplication node to form $\widehat{\mathbf S}$. Its spatial sum gives $\widehat N$, while $\mathcal U_{mp}$ produces the density map at image resolution used for display. The Audit head is an auxiliary output and does not return to the counting path.}
    \label{fig:overall_architecture}
\end{figure*}

Fig.~\ref{fig:overall_architecture} gives the complete forward path. A residual backbone and a feature pyramid network (FPN) first convert $\mathbf I$ into a shared feature tensor $\mathbf F$. An FPN combines features from shallow and deep stages so that the final tensor contains both spatial detail and a larger image context. A fixed image rule also produces an effective valid mask $\mathbf M^{eff}$, which removes nearly black padding and its boundary.

Candidate Evidence (CE) uses $\mathbf F$ to form a broad nonnegative response map $\mathbf C^{all}$. Candidate Acceptance (CA) produces a gate $\mathbf A\in[0,1]$ that favors compact responses near the center patterns learned from source points. Bias Isolation (BI) produces a retention gate $\mathbf V\in[0,1]$. A small value of $\mathbf V$ reduces a response that agrees with repeated lines, broad responses away from a center, grids, or the learned source background response. The final density map on the network output grid is
\begin{equation}
    \widehat{\mathbf S}(\mathbf u)
    =
    \mathbf C^{all}(\mathbf u)
    \mathbf A(\mathbf u)
    \mathbf V(\mathbf u)
    \mathbf M^{eff}(\mathbf u).
    \label{eq:cobicount_factorization_revised}
\end{equation}
Here, $\mathbf u$ is one location on the network output grid. Equation~\eqref{eq:cobicount_factorization_revised} is the main forward relation: a location contributes strongly only when CE supplies a response, CA accepts it, BI retains it, and the location is valid.

The network stride is $\rho=4$. For the image sizes used in this study, the grid of $\widehat{\mathbf S}$ has height $h=H/4$ and width $w=W/4$:
\begin{equation}
    \Omega_{\mathbf S}
    =
    \{0,\ldots,w-1\}
    \times
    \{0,\ldots,h-1\}.
    \label{eq:score_grid_definition}
\end{equation}
Thus, one step on this grid corresponds to four input pixels. The numerical count is obtained before any display resizing:
\begin{equation}
    \widehat N
    =
    \sum_{\mathbf u\in\Omega_{\mathbf S}}
    \widehat{\mathbf S}(\mathbf u).
    \label{eq:count_prediction_method}
\end{equation}
For visualization at image resolution, bilinear resizing is followed by an area correction:
\begin{equation}
    \widehat{\mathbf D}_{\mathbf I}
    =
    \mathcal U_{mp}(\widehat{\mathbf S})
    =
    \frac{hw}{HW}
    \mathcal U_{bil}(\widehat{\mathbf S};H,W),
    \label{eq:mass_preserving_resize}
\end{equation}
where $\mathcal U_{bil}$ denotes bilinear resizing. The factor $hw/(HW)$ compensates for the change in the number of spatial values. The count in Eq.~\eqref{eq:count_prediction_method} is always computed from $\widehat{\mathbf S}$, so it is not affected by interpolation. For numerical safety, the implementation limits each final value and the summed count to $10^{5}$; this limit is not reached in the reported experiments.

The Audit head is separate from the counting path. It receives $\mathbf F$ and selected internal maps, then produces four response maps for later inspection. It has no arrow to $\widehat{\mathbf S}$ or $\widehat N$.

\subsection{Shared Feature and Effective Valid Mask}
\label{subsec:shared_feature_mask}

The residual backbone + FPN begins with a convolutional stem and then uses four residual stages. A residual block adds its transformed feature to a direct or projected copy of its input. This addition helps information pass through the network. The four stages produce features at strides $4$, $8$, $16$, and $32$. Their channel numbers are $64$, $128$, $192$, and $256$, respectively. The FPN applies a $1\times1$ lateral convolution at each stage and passes deeper information toward finer resolutions. It outputs four tensors $\{\mathbf P_2,\mathbf P_3,\mathbf P_4,\mathbf P_5\}$, each with $96$ channels.

The last three tensors are resized to the resolution of $\mathbf P_2$ and concatenated with it:
\begin{equation}
    \mathbf F^{cat}
    =
    [
    \mathbf P_2,
    \mathcal U(\mathbf P_3),
    \mathcal U(\mathbf P_4),
    \mathcal U(\mathbf P_5)
    ],
    \label{eq:feature_concat}
\end{equation}
where $\mathcal U$ denotes bilinear resizing to the size of $\mathbf P_2$. Two $3\times3$ convolution blocks fuse the concatenated tensor:
\begin{equation}
    \mathbf F
    =
    \phi_{fuse}(\mathbf F^{cat})
    \in
    \mathbb R^{96\times h\times w}.
    \label{eq:shared_feature}
\end{equation}
Each block contains a convolution, batch normalization, and a sigmoid linear unit (SiLU). Batch normalization rescales intermediate channels during training, and SiLU is a smooth activation function. Unless stated otherwise, a prediction head $h_z$ used below has a $1\times1$ convolution from $96$ to $48$ channels, a SiLU activation, and a second $1\times1$ convolution that produces the required output channels.

The valid mask is obtained directly from the image and has no learned parameter. Before entering the network, each image channel is normalized using the mean $(0.485,0.456,0.406)$ and standard deviation $(0.229,0.224,0.225)$ commonly used for ImageNet. To construct the valid mask, this numerical transformation is reversed to recover the RGB values. The recovered values are limited to $[0,1]$, and the three channels are averaged to obtain the intensity map $\mathbf J$. The initial valid mask is
\begin{equation}
    \mathbf M^{valid}
    =
    \mathbf 1[\mathbf J>0.035].
    \label{eq:valid_mask}
\end{equation}
where $\mathbf 1[\cdot]$ equals one when its condition is true and zero otherwise.
The complement $1-\mathbf M^{valid}$ marks nearly black pixels. A $9\times9$ maximum pooling operation expands this area to obtain the neutral mask
\begin{equation}
    \mathbf M^{neu}
    =
    \operatorname{MaxPool}_{9\times9}
    (1-\mathbf M^{valid}).
    \label{eq:neutral_mask}
\end{equation}
Maximum pooling keeps the largest value in each neighborhood, so this operation adds a narrow buffer around the black area. Both masks are resized to $h\times w$ with nearest neighbor interpolation. After resizing, the same symbols denote the masks on the network output grid. The effective valid mask is
\begin{equation}
    \mathbf M^{eff}
    =
    \mathbf M^{valid}
    \odot
    (1-\mathbf M^{neu}),
    \label{eq:effective_valid_mask}
\end{equation}
where $\odot$ denotes elementwise multiplication. This mask removes padding; it does not decide whether a valid image location is an object or background. The same fixed rule is used for source images and target images.

\subsection{Candidate Evidence}
\label{subsec:method_ce}

\begin{figure}[!t]
    \centering
    \includegraphics[width=\columnwidth]{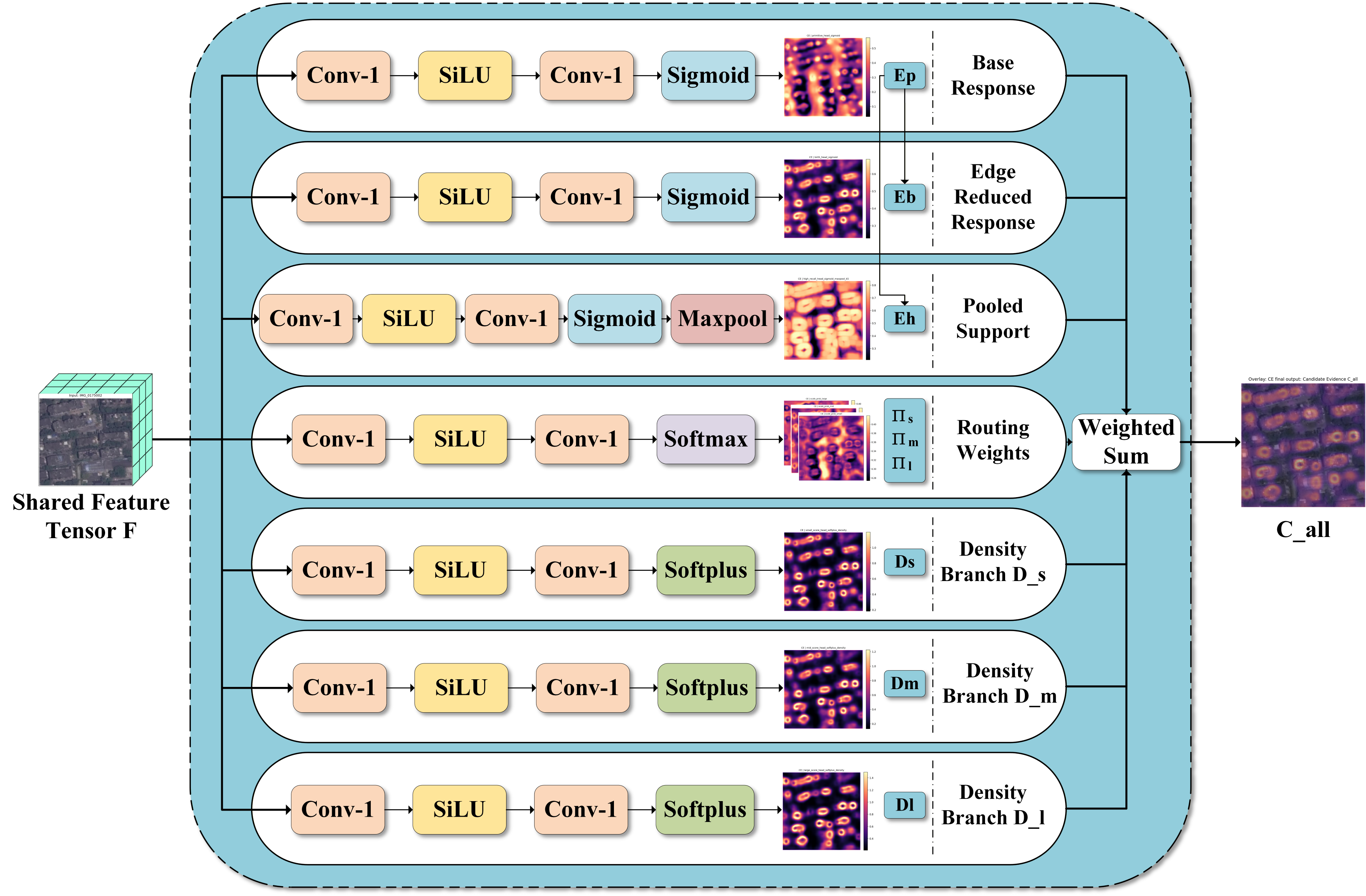}
    \caption{Candidate Evidence. The shared feature tensor first produces $\mathbf E_p$. Its Sobel response $\mathbf G$ enters the computation of $\mathbf E_b$, and $\mathbf E_p$ enters the computation of $\mathbf E_h$. The scale head supplies $\pi_s$, $\pi_m$, and $\pi_l$, which multiply the matching density head outputs. The weighted sum in Eq.~\eqref{eq:candidate_evidence} combines $\mathbf D^{ms}$, $\mathbf E_p$, $\mathbf E_b$, and $\mathbf E_h$ into $\mathbf C^{all}$.}
    \label{fig:ce_module}
\end{figure}

CE performs the first decision in the counting path: it creates possible responses before CA and BI remove unsupported ones. The module uses three bounded response maps and three routed density maps. The routed density $\mathbf D^{ms}$ is the contribution from the three nonnegative density branches. By contrast, $\mathbf E_p$, $\mathbf E_b$, and $\mathbf E_h$ describe the base response, the response after strong edges are reduced, and the response supported by a wider neighborhood. These four terms are combined into one CE map; none is treated as a separate final prediction. Fig.~\ref{fig:ce_module} summarizes these computations, while Eqs.~\eqref{eq:primitive_evidence_revised}--\eqref{eq:candidate_evidence} give their exact inputs and order.

The base response is
\begin{equation}
    \mathbf E_p
    =
    \sigma(h_p(\mathbf F))
    \odot
    \mathbf M^{eff},
    \label{eq:primitive_evidence_revised}
\end{equation}
where $\sigma$ is the sigmoid function, which limits each value to $[0,1]$. The change around this response is measured by a Sobel filter. The two fixed kernels are
\begin{equation}
    S_x
    =
    \frac{1}{8}
    \begin{bmatrix}
    -1&0&1\\
    -2&0&2\\
    -1&0&1
    \end{bmatrix},
    \qquad
    S_y=S_x^{\mathsf T}.
    \label{eq:sobel_kernels}
\end{equation}
For any one channel map $\mathbf X$, the following operation rescales its values to $[0,1]$ within each image:
\begin{equation}
\begin{aligned}
    \operatorname{Norm}_{01}(\mathbf X)
    &=
    \frac{
    \mathbf X-\min_{\mathbf u}\mathbf X(\mathbf u)
    }{
    \max_{\mathbf u}\mathbf X(\mathbf u)
    -\min_{\mathbf u}\mathbf X(\mathbf u)
    +\epsilon_{norm}
    },\\
    \epsilon_{norm}&=10^{-6}.
\end{aligned}
    \label{eq:norm01_definition}
\end{equation}
The normalized Sobel magnitude is
\begin{equation}
    \mathbf G
    =
    \operatorname{Norm}_{01}
    \left(
    \sqrt{
    (S_x*\mathbf E_p)^2
    +(S_y*\mathbf E_p)^2
    +10^{-6}
    }
    \right),
    \label{eq:sobel_response}
\end{equation}
where $*$ denotes convolution. A large $\mathbf G$ indicates a strong local change in $\mathbf E_p$.

The second response reduces values at these strong changes:
\begin{equation}
    \mathbf E_b
    =
    \sigma(h_b(\mathbf F))
    \odot
    \operatorname{clip}(1-0.35\mathbf G,0,1)
    \odot
    \mathbf M^{eff}.
    \label{eq:birth_evidence_revised}
\end{equation}
The function $\operatorname{clip}(\mathbf X,a,b)$ limits every value of $\mathbf X$ to $[a,b]$. The third response enlarges nearby support:
\begin{equation}
\begin{aligned}
    \mathbf E_h
    &=
    \operatorname{MaxPool}_{5\times5}
    \Big[
    \sigma(h_h(\mathbf F))\\
    &\quad\odot
    (0.5+0.5\mathbf E_p)
    \Big]
    \odot
    \mathbf M^{eff}.
\end{aligned}
    \label{eq:high_recall_revised}
\end{equation}
The factor $(0.5+0.5\mathbf E_p)$ ties this response to the base response. Maximum pooling then extends it over a $5\times5$ neighborhood.

The remaining CE path allows three density heads to contribute by different amounts at each location. The scale head produces
\begin{equation}
\begin{aligned}
    \bigl[\pi_s,\pi_m,\pi_l\bigr]
    &=
    \operatorname{Softmax}_{ch}
    (h_{scale}(\mathbf F)),\\
    \pi_s+\pi_m+\pi_l&=1.
\end{aligned}
    \label{eq:scale_prob}
\end{equation}
Softmax is applied across the three channels, so the three values form local routing weights. Three other heads produce nonnegative maps:
\begin{equation}
    \mathbf D_k
    =
    \operatorname{Softplus}(h_k(\mathbf F))
    \odot
    \mathbf M^{eff},
    \qquad
    k\in\{s,m,l\}.
    \label{eq:scale_density_heads}
\end{equation}
Softplus converts an unrestricted head output into a nonnegative value. The labels $s$, $m$, and $l$ are only branch identifiers. No object size label is supplied, and the method does not assume that these branches have a fixed physical scale. Their routed sum is
\begin{equation}
    \mathbf D^{ms}
    =
    \pi_s\odot\mathbf D_s
    +
    \pi_m\odot\mathbf D_m
    +
    \pi_l\odot\mathbf D_l.
    \label{eq:multi_scale_density}
\end{equation}
The superscript $ms$ denotes this mixture of three routed branches; it is not a separate prediction head.

The CE output combines the routed density and the three bounded responses:
\begin{equation}
    \mathbf C^{all}
    =
    0.34\mathbf D^{ms}
    +0.24\mathbf E_p
    +0.22\mathbf E_b
    +0.20\mathbf E_h.
    \label{eq:candidate_evidence}
\end{equation}
The coefficients are fixed implementation values. They sum to one, but $\mathbf C^{all}$ is not a probability map and its spatial sum is not constrained to equal one.

\subsection{Candidate Acceptance}
\label{subsec:method_ca}

\begin{figure}[!t]
    \centering
    \includegraphics[width=0.98\columnwidth]{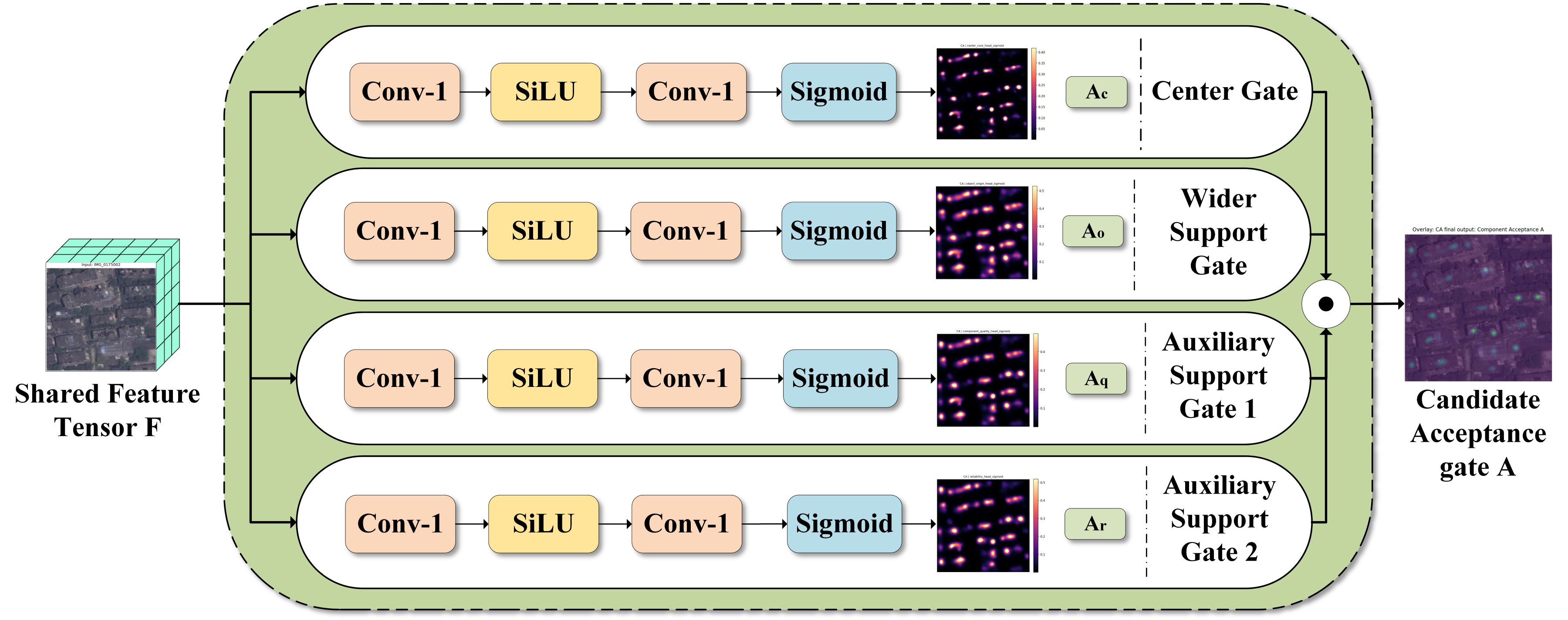}
    \caption{Candidate Acceptance. Four heads produce $\mathbf A_c$, $\mathbf A_o$, $\mathbf A_q$, and $\mathbf A_r$. The last two maps form $0.35+0.65\mathbf A_q$ and $0.35+0.65\mathbf A_r$, respectively, before the four factors enter one multiplication node. The output is the acceptance gate $\mathbf A$.}
    \label{fig:ca_module}
\end{figure}

CE is deliberately broad, so one object response may spread over several locations or contain several nearby peaks. CA performs the second decision: it controls how much of each CE response can continue to the final density map. It predicts four maps from $\mathbf F$:
\begin{equation}
\begin{aligned}
    \mathbf A_c&=\sigma(h_c(\mathbf F)),
    &\mathbf A_o&=\sigma(h_o(\mathbf F)),\\
    \mathbf A_q&=\sigma(h_q(\mathbf F)),
    &\mathbf A_r&=\sigma(h_r(\mathbf F)).
\end{aligned}
    \label{eq:acceptance_cues}
\end{equation}
$\mathbf A_c$ is trained to be high near a source point center. $\mathbf A_o$ is trained with a wider area around each source point. The two remaining maps, $\mathbf A_q$ and $\mathbf A_r$, use the same wider target with different loss weights. The implementation names them quality and reliability, but they do not represent two separately annotated properties.

\begin{figure*}[!t]
    \centering
    \includegraphics[width=0.90\textwidth]{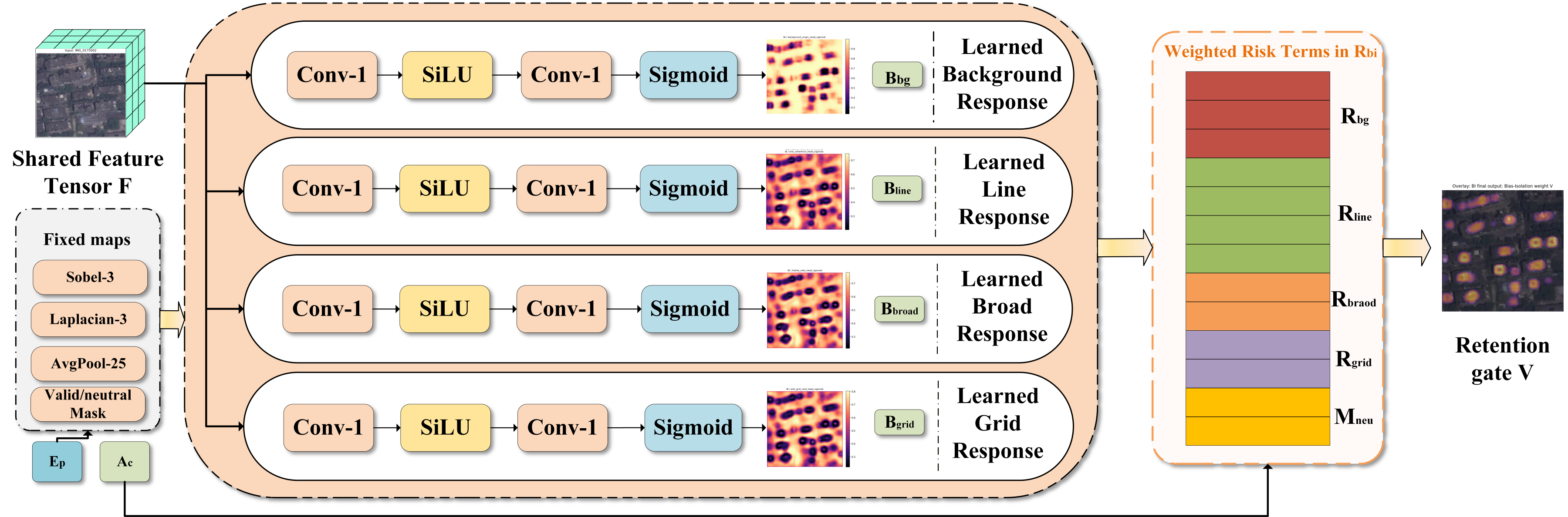}
    \caption{Bias Isolation. The exact inputs are $\mathbf F$, $\mathbf E_p$, $\mathbf A_c$, and $\mathbf M^{neu}$. Fixed filters applied to $\mathbf E_p$ produce $\mathbf G$, $\mathbf L$, and $\mathbf P$. These maps combine with four learned maps to form $\mathbf R_{line}$, $\mathbf R_{broad}$, $\mathbf R_{grid}$, and $\mathbf R_{bg}$. The weighted sum in Eq.~\eqref{eq:bias_risk_exact} produces $\mathbf R^{bi}$, followed by $\operatorname{clip}(1-\mathbf R^{bi},0,1)$ to obtain $\mathbf V$.}
    \label{fig:bi_module}
\end{figure*}

The four maps form
\begin{equation}
\begin{aligned}
    \mathbf A
    =
    &\mathbf A_c
    \odot
    \mathbf A_o
    \odot
    (0.35+0.65\mathbf A_q)\\
    &\odot
    (0.35+0.65\mathbf A_r).
\end{aligned}
    \label{eq:component_acceptance}
\end{equation}
The first two maps can strongly reduce a response. Each of the last two factors remains between $0.35$ and $1$, so neither can remove a response by itself. Here, a compact response does not mean that CA measures its shape. Instead, the narrow center target and the wider support target encourage an accepted response to remain near a source point and to decrease away from it. CA does not find connected components or apply a hand written shape rule. It learns these four spatial gates from source point supervision.

\subsection{Bias Isolation}
\label{subsec:method_bi}

BI performs the third decision. Here, bias means a repeated source image pattern that can trigger a counting response even when no annotated object supports it. BI does not recognize named background categories in a target domain. Instead, it estimates four continuous risk maps and converts their weighted sum into the retention gate $\mathbf V$.

Four learned maps are obtained from $\mathbf F$:
\begin{equation}
\begin{aligned}
    \mathbf B_{bg}&=\sigma(h_{bg}(\mathbf F)),
    &\mathbf B_{line}&=\sigma(h_{line}(\mathbf F)),\\
    \mathbf B_{broad}&=\sigma(h_{broad}(\mathbf F)),
    &\mathbf B_{grid}&=\sigma(h_{grid}(\mathbf F)).
\end{aligned}
    \label{eq:bi_learned_maps}
\end{equation}
$\mathbf B_{bg}$ is the only map in this group with a direct background target. The other three maps receive no labels for roads, lines, parking grids, or other named structures, so they are not classifiers of named structures. Their names refer to the response structures used in Eqs.~\eqref{eq:line_risk}--\eqref{eq:grid_risk}. They receive gradients through the retention gate $\mathbf V$ and the final density map $\widehat{\mathbf S}$. The source density and count losses keep counting mass near annotated source objects, while the suppression losses penalize remaining mass outside the protected source regions. The three maps are therefore learned from source counting supervision rather than from named structure labels.

BI also computes fixed transforms from the base response $\mathbf E_p$. The Sobel map $\mathbf G$ was defined in Eq.~\eqref{eq:sobel_response}. The four neighbor Laplacian kernel is
\begin{equation}
    K_{lap}
    =
    \begin{bmatrix}
    0&1&0\\
    1&-4&1\\
    0&1&0
    \end{bmatrix}.
    \label{eq:laplacian_kernel}
\end{equation}
The Laplacian measures local second order change. Average pooling measures the mean response in a larger neighborhood. Their normalized maps are
\begin{equation}
\begin{aligned}
    \mathbf L
    &=
    \operatorname{Norm}_{01}
    (|K_{lap}*\mathbf E_p|),\\
    \mathbf P
    &=
    \operatorname{Norm}_{01}
    (\operatorname{AvgPool}_{25\times25}(\mathbf E_p)).
\end{aligned}
    \label{eq:bi_fixed_maps}
\end{equation}
$\mathbf L$ becomes large where the response changes repeatedly over a short distance. $\mathbf P$ becomes large where a response remains broad over a larger area.

The learned and fixed maps are combined in a fixed order:
\begin{align}
    \mathbf R_{line}
    &=
    \operatorname{clip}
    (0.55\mathbf B_{line}+0.45\mathbf G,0,1),
    \label{eq:line_risk}\\
    \mathbf R_{broad}
    &=
    \operatorname{clip}
    \left(
    0.60\mathbf B_{broad}
    +0.40\mathbf P\odot(1-\mathbf A_c),
    0,1
    \right),
    \label{eq:broad_risk}\\
    \mathbf R_{grid}
    &=
    \operatorname{clip}
    \left(
    0.65\mathbf B_{grid}
    +0.35\mathbf L\odot(1-\mathbf A_c),
    0,1
    \right).
    \label{eq:grid_risk}
\end{align}
$\mathbf R_{line}$ responds to a learned line map and a strong first order change. $\mathbf R_{broad}$ becomes large when the pooled response is strong but the center gate is weak. It does not identify enclosed empty regions as a separate class. $\mathbf R_{grid}$ combines repeated second order changes with a weak center gate. It is not trained with parking grid labels.

The background risk also uses the neutral mask:
\begin{equation}
\begin{aligned}
    \mathbf R_{bg}
    &=
    \operatorname{clip}
    \Bigl(
    \mathbf B_{bg}
    \odot
    (0.50+0.50\mathbf R_{broad})\\
    &\qquad
    +0.25\mathbf M^{neu},0,1
    \Bigr).
\end{aligned}
    \label{eq:background_like_exact}
\end{equation}
The four risk maps may overlap; they are not four exclusive classes. Their weighted sum is
\begin{equation}
\begin{aligned}
    \mathbf R^{bi}
    =
    &0.72\mathbf R_{bg}
    +0.58\mathbf R_{line}
    +0.48\mathbf R_{broad}\\
    &+0.60\mathbf R_{grid}
    +1.00\mathbf M^{neu}.
\end{aligned}
    \label{eq:bias_risk_exact}
\end{equation}
Finally,
\begin{equation}
    \mathbf V
    =
    \operatorname{clip}(1-\mathbf R^{bi},0,1).
    \label{eq:veto_exact}
\end{equation}
A large risk therefore produces a small retention value at the same location. The neutral mask enters both $\mathbf R_{bg}$ and $\mathbf R^{bi}$, which strongly reduces responses near black padding.

\subsection{Audit Response Maps}
\label{subsec:audit_maps}

Audit is an auxiliary head for inspecting response patterns. It does not change $\mathbf C^{all}$, $\mathbf A$, $\mathbf V$, $\widehat{\mathbf S}$, or $\widehat N$. A four channel head first produces the raw maps
\begin{equation}
    [
    \mathbf Z^0_{veh},
    \mathbf Z^0_{build},
    \mathbf Z^0_{ship},
    \mathbf Z^0_{bg}
    ]
    =
    h_{aud}(\mathbf F).
    \label{eq:audit_base_logits}
\end{equation}
The channel names vehicle, building, ship, and background describe response patterns for inspection; they are not predicted object classes. Fixed offsets then connect each raw map to selected internal responses:
\begin{align}
    \mathbf Z_{veh}
    &=
    \mathbf Z^0_{veh}
    +0.75\pi_s
    +0.20\pi_m
    +0.20\mathbf A_c,
    \label{eq:audit_vehicle_logit}\\
    \mathbf Z_{build}
    &=
    \mathbf Z^0_{build}
    +0.80\pi_l
    +0.35\mathbf R_{broad},
    \label{eq:audit_building_logit}\\
    \mathbf Z_{ship}
    &=
    \mathbf Z^0_{ship}
    +0.50\mathbf R_{line}
    +0.20\pi_m,
    \label{eq:audit_ship_logit}\\
    \mathbf Z_{bg}
    &=
    \mathbf Z^0_{bg}
    +1.10\mathbf R_{bg}
    +0.35\mathbf R_{grid}.
    \label{eq:audit_background_logit}
\end{align}
These fixed links are inspection rules. They do not mean that $\pi_s$ denotes vehicles, $\pi_l$ denotes buildings, or $\mathbf R_{line}$ denotes ships. More generally, a routing branch or a structural risk is not assigned to an unseen target class.

The four Audit response maps are
\begin{equation}
\begin{aligned}
    \mathbf P^{aud}
    &=
    [
    \mathbf P^{aud}_{veh},
    \mathbf P^{aud}_{build},
    \mathbf P^{aud}_{ship},
    \mathbf P^{aud}_{bg}
    ]\\
    &=
    \operatorname{Softmax}_{ch}
    ([
    \mathbf Z_{veh},
    \mathbf Z_{build},
    \mathbf Z_{ship},
    \mathbf Z_{bg}
    ]).
\end{aligned}
    \label{eq:audit_probabilities}
\end{equation}
These values sum to one across channels at each location. Their numerical values should not be read as reliable probabilities of unseen target classes. In the main protocol, only the building and background channels receive direct labels from the RSOC Building source. The other channels obtain no vehicle or ship label during training. Audit should therefore be read only as a visual description of internal response patterns.

\subsection{Training with Source Points}
\label{subsec:method_training_objective}

All training maps and losses are computed from a source image and its point set. They are introduced below in the same order in which they are used: first the training maps constructed from source points, then the density and count losses, then the losses for CA and BI, and finally the routing and Audit terms.

\subsubsection{Maps constructed from source points}

For the following construction on one source image, we omit the image index used in Section~\ref{subsec:method_problem} and write its annotated point set as
$\mathcal P^s=\{\mathbf p_i\}_{i=1}^{N^*}$,
where $N^*=|\mathcal P^s|$ is the reference count for this image. Because the network output grid has stride $\rho=4$, each point coordinate is divided by $\rho$ and rounded to the nearest grid location:
\begin{equation}
    \widetilde{\mathbf p}_i
    =
    \operatorname{round}
    \left(
    \frac{\mathbf p_i}{\rho}
    \right),
    \qquad
    \rho=4.
    \label{eq:source_point_projection}
\end{equation}
Both coordinates are transformed in this operation.

Each source annotation provides only an object location. To construct the density target used for training, a normalized Gaussian kernel is placed around each projected point. For a Gaussian width $\sigma$, its finite window is
\begin{equation}
\begin{aligned}
    \mathcal W_{\sigma}
    &=
    \left\{
    \boldsymbol\delta=(\delta_x,\delta_y):
    |\delta_x|,|\delta_y|\le r_{\sigma}
    \right\},\\
    r_{\sigma}
    &=
    \max(1,\operatorname{round}(3\sigma)).
\end{aligned}
    \label{eq:gaussian_window}
\end{equation}
The discrete kernel inside this window has unit sum:
\begin{equation}
    \kappa_{\sigma}(\boldsymbol\delta)
    =
    \frac{
    \exp(-\|\boldsymbol\delta\|_2^2/(2\sigma^2))
    }{
    \sum_{\boldsymbol\delta'\in\mathcal W_{\sigma}}
    \exp(-\|\boldsymbol\delta'\|_2^2/(2\sigma^2))
    }.
    \label{eq:discrete_gaussian_kernel}
\end{equation}
One kernel is added at each projected point to form the source density target:
\begin{equation}
    \mathbf D^*(\mathbf u)
    =
    \sum_{i=1}^{N^*}
    \kappa_{\sigma_p}
    (\mathbf u-\widetilde{\mathbf p}_i),
    \qquad
    \sigma_p=1.15.
    \label{eq:source_density_target}
\end{equation}
Values outside $\Omega_{\mathbf S}$ are discarded. The kernel is normalized before it is placed, but a kernel cut by an image boundary is not normalized again. Thus, $\mathbf D^*$ is a density target constructed from source annotations rather than another model output. During training, it is compared with the predicted density map $\widehat{\mathbf S}$ in Eq.~\eqref{eq:loss_density}. It is not used when the fixed model processes a target image.

CA needs a center target and a wider support target. These maps use Gaussians with a maximum value of one rather than a unit sum:
\begin{equation}
\begin{aligned}
    \mathbf T_{\sigma}(\mathbf u)
    =
    \max_{1\le i\le N^*}
    \Bigg[
    &\exp\left(
    -\frac{
    \|\mathbf u-\widetilde{\mathbf p}_i\|_2^2
    }{2\sigma^2}
    \right)\\
    &\cdot
    \mathbf 1[
    \mathbf u-\widetilde{\mathbf p}_i
    \in\mathcal W_{\sigma}
    ]
    \Bigg].
\end{aligned}
    \label{eq:peak_normalized_gaussian_map}
\end{equation}
When no point is present, $\mathbf D^*$ and $\mathbf T_{\sigma}$ are zero maps. The center target and support target are
\begin{equation}
    \mathbf T^{peak}=\mathbf T_{\sigma_p},
    \qquad
    \mathbf T^{sup}=\mathbf T_{\sigma_s},
    \qquad
    \sigma_s=2.35.
    \label{eq:center_support_targets}
\end{equation}
$\mathbf T^{peak}$ is narrow and $\mathbf T^{sup}$ covers a wider area. Their binary masks are
\begin{equation}
\begin{aligned}
    \mathbf M^{pt}
    &=
    \mathbf 1[\mathbf T^{peak}>0.05],\\
    \mathbf M^{sup}
    &=
    \mathbf 1[\mathbf T^{sup}>0.05],\\
    \mathbf M^{fg}
    &=
    \operatorname{Dilate}_{9\times9}
    (\mathbf M^{pt}),\\
    \mathbf M^{bg}
    &=
    (1-\mathbf M^{fg})
    \odot
    \mathbf M^{eff}.
\end{aligned}
    \label{eq:source_support_masks}
\end{equation}
$\operatorname{Dilate}_{9\times9}$ is binary maximum pooling. $\mathbf M^{sup}$ marks the area used by the support losses. $\mathbf M^{fg}$ is a separate protection area used by the background and strong response losses. $\mathbf M^{bg}$ contains valid locations outside $\mathbf M^{fg}$ and supplies the direct target for $\mathbf B_{bg}$.

\subsubsection{Density and count losses}

For compact notation, $\langle\mathbf X\rangle$ denotes the mean over all spatial locations and all images in the current batch. The two elementary penalties are
\begin{equation}
    \ell_{SL1}(z)
    =
    \begin{cases}
        \frac{1}{2}z^2, & |z|<1,\\
        |z|-\frac{1}{2}, & |z|\ge1,
    \end{cases}
    \label{eq:smooth_l1_definition}
\end{equation}
and
\begin{equation}
    \ell_{BCE}(p,y)
    =
    -y\log p-(1-y)\log(1-p).
    \label{eq:bce_definition}
\end{equation}
The first is the smooth $L_1$ penalty. The second is binary cross entropy for a bounded prediction $p$ and target $y$.
Before binary cross entropy is evaluated, the implementation limits $p$ away from zero and one by $10^{-4}$ for numerical stability.

Let $N_{norm}=\max(N^*,1)$. The density loss is
\begin{equation}
\begin{aligned}
    \mathcal L_{den}
    =
    \Bigg\langle
    &\ell_{SL1}
    \left(
    \frac{\widehat{\mathbf S}}{N_{norm}}
    -
    \frac{\mathbf D^*}{N_{norm}}
    \right)\\
    &\odot
    (1+4\mathbf M^{sup})
    \odot
    \mathbf M^{eff}
    \Bigg\rangle.
\end{aligned}
    \label{eq:loss_density}
\end{equation}
Division by $N_{norm}$ prevents images with many objects from dominating this map loss. The factor $(1+4\mathbf M^{sup})$ gives more weight to locations near source objects.

Two losses compare the summed prediction with the reference count. They are evaluated for each image and then averaged over the batch:
\begin{align}
    \mathcal L_{cnt}^{rel}
    &=
    \frac{|\widehat N-N^*|}{N^*+1},
    \label{eq:loss_count_relative}\\
    \mathcal L_{cnt}^{log}
    &=
    |\log(1+\widehat N)-\log(1+N^*)|.
    \label{eq:loss_count}
\end{align}
The relative term limits the effect of large reference counts. The logarithmic term reduces the difference between very large numerical ranges.

\subsubsection{Losses for CA and BI}

The CA maps are trained with the center and support targets defined above. Their spatial weights are
\begin{equation}
\begin{aligned}
    \boldsymbol\omega_c
    &=
    (1+7\mathbf T^{peak})
    \odot
    \mathbf M^{eff},\\
    \boldsymbol\omega_o
    &=
    (1+4\mathbf M^{sup})
    \odot
    \mathbf M^{eff},\\
    \boldsymbol\omega_s
    &=
    (1+2\mathbf M^{sup})
    \odot
    \mathbf M^{eff}.
\end{aligned}
    \label{eq:acceptance_loss_weights}
\end{equation}
The corresponding losses are
\begin{align}
    \mathcal L_{ctr}
    &=
    \left\langle
    \boldsymbol\omega_c
    \odot
    \ell_{BCE}(\mathbf A_c,\mathbf T^{peak})
    \right\rangle,
    \label{eq:loss_center}\\
    \mathcal L_{obj}
    &=
    \left\langle
    \boldsymbol\omega_o
    \odot
    \ell_{BCE}(\mathbf A_o,\mathbf T^{sup})
    \right\rangle,
    \label{eq:loss_object_support}\\
    \mathcal L_{bg}
    &=
    \left\langle
    \mathbf M^{eff}
    \odot
    \ell_{BCE}(\mathbf B_{bg},\mathbf M^{bg})
    \right\rangle,
    \label{eq:loss_background_head}\\
    \mathcal L_{org}
    &=
    \mathcal L_{obj}+0.55\mathcal L_{bg},
    \label{eq:loss_origin}\\
    \mathcal L_{rly}
    &=
    \left\langle
    \boldsymbol\omega_s
    \odot
    \ell_{BCE}(\mathbf A_r,\mathbf T^{sup})
    \right\rangle,
    \label{eq:loss_reliability}\\
    \mathcal L_{qua}
    &=
    \left\langle
    \boldsymbol\omega_s
    \odot
    \ell_{BCE}(\mathbf A_q,\mathbf T^{sup})
    \right\rangle.
    \label{eq:loss_ca}
\end{align}
These losses state where CA should remain open and where the learned background response should be high. They do not provide labels for any named target background structure.

BI is also trained through penalties on the final counting mass. Let $\operatorname{sg}(\cdot)$ denote stop gradient. It keeps a map's value in the forward calculation but blocks the direct gradient through that map when it acts as a weight. The four penalties are
\begin{align}
    \mathcal L_{bgv}
    &=
    \left\langle
    \widehat{\mathbf S}
    \odot
    \operatorname{sg}(\mathbf R_{bg})
    \odot
    (1-\mathbf M^{fg})
    \odot
    \mathbf M^{eff}
    \right\rangle,
    \label{eq:loss_background_risk}\\
    \mathcal L_{line}
    &=
    \left\langle
    \widehat{\mathbf S}
    \odot
    \operatorname{sg}(\mathbf R_{line})
    \odot
    (1-\mathbf M^{sup})
    \odot
    \mathbf M^{eff}
    \right\rangle,
    \label{eq:loss_line_risk}\\
    \mathcal L_{broad}
    &=
    \left\langle
    \widehat{\mathbf S}
    \odot
    \operatorname{sg}(\mathbf R_{broad})
    \odot
    (1-\mathbf M^{sup})
    \odot
    \mathbf M^{eff}
    \right\rangle,
    \label{eq:loss_broad_risk}\\
    \mathcal L_{grid}
    &=
    \left\langle
    \widehat{\mathbf S}
    \odot
    \operatorname{sg}(\mathbf R_{grid})
    \odot
    (1-\mathbf M^{sup})
    \odot
    \mathbf M^{eff}
    \right\rangle.
    \label{eq:loss_bi}
\end{align}
The background penalty uses the wider protection mask $\mathbf M^{fg}$. The other three use $\mathbf M^{sup}$. Although the risk maps are detached in these four weighting positions, their heads still receive gradients through $\mathbf V$ and the final density map $\widehat{\mathbf S}$.

The strongest remaining responses outside $\mathbf M^{fg}$ receive an additional penalty. These locations are called hard negatives because source point supervision treats them as background, but the current model still gives them a large response. For each image, define
\begin{equation}
    \mathbf Q
    =
    \operatorname{sg}(\widehat{\mathbf S})
    \odot
    (1-\mathbf M^{fg})
    \odot
    \mathbf M^{eff}.
    \label{eq:hard_negative_selector}
\end{equation}
Let $\mathcal H_{K_{hn}}$ contain the $K_{hn}$ largest locations of $\mathbf Q$, where
\begin{equation}
    K_{hn}
    =
    \min(768,|\Omega_{\mathbf S}|).
    \label{eq:hard_negative_number}
\end{equation}
The loss on these locations is
\begin{equation}
    \mathcal L_{hn}
    =
    \frac{1}{|\mathcal H_{K_{hn}}|}
    \sum_{\mathbf u\in\mathcal H_{K_{hn}}}
    \widehat{\mathbf S}(\mathbf u)
    (1-\mathbf M^{fg}(\mathbf u))
    \mathbf M^{eff}(\mathbf u).
    \label{eq:loss_hard_negative}
\end{equation}
$\operatorname{sg}(\widehat{\mathbf S})$ is used only to select the locations. The selected values of $\widehat{\mathbf S}$ in Eq.~\eqref{eq:loss_hard_negative} remain differentiable.

\subsubsection{Routing, Audit, and total loss}

Two terms prevent the routing weights from assigning nearly all local weight to a single branch. Their local entropy is
\begin{equation}
    \mathcal H_{\pi}(\mathbf u)
    =
    -\sum_{k\in\{s,m,l\}}
    \pi_k(\mathbf u)\log\pi_k(\mathbf u).
    \label{eq:scale_entropy_definition}
\end{equation}
The entropy loss is
\begin{equation}
    \mathcal L_{ent}
    =
    \left\langle
    (1.10-\mathcal H_{\pi})
    \odot
    (1-\mathbf M^{sup})
    \odot
    \mathbf M^{eff}
    \right\rangle.
    \label{eq:loss_scale_entropy}
\end{equation}
Entropy is large when the three weights are similar and small when one weight dominates. The value $1.10$ approximates $\log 3$, the largest entropy of three routing weights. Minimizing $\mathcal L_{ent}$ avoids an early, highly concentrated choice outside the support area.

Let $B$ be the batch size. The mean use of branch $k$ is
\begin{equation}
    \overline\pi_k
    =
    \frac{1}{B|\Omega_{\mathbf S}|}
    \sum_{b=1}^{B}
    \sum_{\mathbf u\in\Omega_{\mathbf S}}
    \pi_{b,k}(\mathbf u).
    \label{eq:batch_scale_average}
\end{equation}
The balance loss is
\begin{equation}
\begin{aligned}
    \mathcal L_{bal}
    &=
    \frac{1}{3}
    \sum_{k\in\{s,m,l\}}
    (\overline\pi_k-q_k)^2,\\
    (q_s,q_m,q_l)
    &=
    (0.36,0.34,0.30).
\end{aligned}
    \label{eq:loss_scale_balance}
\end{equation}

For the main protocol, direct Audit supervision begins at epoch $10$. RSOC Building points supervise the building channel inside $\mathbf M^{sup}$, and valid locations outside $\mathbf M^{fg}$ supervise the background channel:
\begin{align}
    \mathcal L_{aud}^{build}
    &=
    -\left\langle
    \log\mathbf P^{aud}_{build}
    \odot
    \mathbf M^{sup}
    \odot
    \mathbf M^{eff}
    \right\rangle,
    \label{eq:loss_audit_building}\\
    \mathcal L_{aud}^{bg}
    &=
    -\left\langle
    \log\mathbf P^{aud}_{bg}
    \odot
    (1-\mathbf M^{fg})
    \odot
    \mathbf M^{eff}
    \right\rangle,
    \label{eq:loss_audit_background}\\
    \mathcal L_{aud}
    &=
    \mathbf 1[e\ge10]
    (\mathcal L_{aud}^{build}+0.35\mathcal L_{aud}^{bg}),
    \label{eq:loss_audit}
\end{align}
where $e$ is the current epoch. The vehicle and ship channels receive no direct source class label.

The complete training objective is
\begin{equation}
\begin{aligned}
    \mathcal L
    =
    &1.25\mathcal L_{den}
    +c_e
    (0.72\mathcal L_{cnt}^{rel}
    +0.20\mathcal L_{cnt}^{log})\\
    &+0.55\mathcal L_{ctr}
    +0.65\mathcal L_{org}
    +0.40\mathcal L_{rly}
    +0.35\mathcal L_{qua}\\
    &+0.38\mathcal L_{bgv}
    +0.22\mathcal L_{line}
    +0.25\mathcal L_{broad}
    +0.30\mathcal L_{grid}\\
    &+0.18\mathcal L_{hn}
    +0.018\mathcal L_{ent}
    +0.020\mathcal L_{bal}
    +0.16\mathcal L_{aud},\\
    &\hspace{17mm}
    c_e=\min\left(1,\frac{e}{8}\right).
\end{aligned}
    \label{eq:total_loss}
\end{equation}
The coefficient $c_e$ increases the count loss during the first eight epochs. CE has no separate direct loss on $\mathbf C^{all}$. It is updated through the final density map, count, BI, strong response, and routing terms. Every quantity in Eq.~\eqref{eq:total_loss} is computed from source images, source points, or fixed transforms of the current source image.

\FloatBarrier
\subsection{Diagnostic Points at Inference}
\label{subsec:diagnostic_interface}

Point extraction is not part of the numerical count. The model first computes $\widehat{\mathbf S}$ and $\widehat N$ by Eqs.~\eqref{eq:cobicount_factorization_revised} and~\eqref{eq:count_prediction_method}. Afterward, a fixed greedy procedure extracts local maxima from $\widehat{\mathbf S}$ only for spatial analysis. The extracted locations are called diagnostic points.

For each image, the selection threshold is
\begin{equation}
    \tau
    =
    \max
    \left(
    0.020,
    0.28
    \max_{\mathbf u}
    \widehat{\mathbf S}(\mathbf u)
    \right).
    \label{eq:diagnostic_threshold}
\end{equation}
The procedure selects the largest remaining value above $\tau$, records its location, and removes its neighborhood before searching again. Accepted points must be at least $3.2$ network output grid cells apart. With stride $4$, this distance is about $12.8$ input pixels. The point budget is
\begin{equation}
    K_{pt}
    =
    \min
    \left(
    1024,
    \max
    [
    1,
    \operatorname{round}(1.55\widehat N+16)
    ]
    \right).
    \label{eq:diagnostic_point_budget}
\end{equation}
This budget only limits the search. It does not force the number of extracted points to equal $\widehat N$. A network output grid point $(x,y)$ is returned to image coordinates by multiplying both coordinates by the stride before visualization or distance evaluation.

For Audit visualization, the four values in $\mathbf P^{aud}$ are sampled at each extracted point. The point receives the label of the largest channel. This label changes neither the point location nor the numerical count. Target annotations are used only by the evaluator to measure the distance between these fixed predictions and reference points.

\section{Experiments}
\label{sec:experiments}

This section first specifies the data access rule, datasets, evaluation measures, comparison settings, and implementation. It then reports the main comparison and tests in which the training source is changed.

\subsection{Protocol and Data}
\label{subsec:exp_protocol_data}

The experiments follow \emph{source-only counting} as defined in Section~\ref{sec:introduction}. The roles of source and target are assigned separately for each experiment; they are not fixed properties of RSOC, DOTA, or DIOR. A source domain contains one object category under a related set of imaging conditions and provides the annotations used to train and select a model. Ordinary training, validation, and test splits of that same category still belong to the same domain. By contrast, a target domain contains a different category or imaging condition that is kept unavailable until the source model has been selected. Thus, a test split from RSOC Building is a source test set when RSOC Building is the training source; it is not treated as a target domain merely because it is used for testing.

In the main experiment, RSOC Building~\cite{gao2020counting} is the single source domain. Only its training split is used to update the model, and only its validation split is used to select the saved model state, hereafter called the checkpoint. After this choice, the model is fixed and evaluated on DOTA Large Vehicle (LV), DOTA Small Vehicle (SV), and DOTA Ship~\cite{xia2018dota}. These three categories are separate unseen target domains in this experiment. This assignment is only one instance of the protocol. Section~\ref{subsec:cross_source_generalization} changes the source to DIOR Airplane~\cite{li2020dior} or DOTA Ship to test whether the results depend on RSOC Building.

The training program reads only the source training and validation splits. The checkpoint with the lowest average absolute count error on source validation is retained. Target images and annotations are read only by the evaluation programs and do not update model parameters or select the checkpoint reported in the main comparison. Target annotations are used solely to compute the final measures.

DOTA represents each object by an oriented quadrilateral. For vertices $\{(x_k,y_k)\}_{k=1}^{4}$, the evaluator uses
\begin{equation}
    \left(
    \frac{1}{4}\sum_{k=1}^{4}x_k,
    \frac{1}{4}\sum_{k=1}^{4}y_k
    \right)
    \label{eq:dota_point_conversion}
\end{equation}
as the reference point, following the supplied parser. DIOR boxes are converted in the same spirit by taking each box center. In both cases, the box or quadrilateral is used only to obtain one point and the object count. Its size, direction, and boundary are not supplied to training or to the extraction of diagnostic points. 

\subsection{Evaluation Measures and Comparison Settings}
\label{subsec:exp_metrics_comparison}

For image $i$, let $N_i^*$ be the reference count and let $\widehat N_i$ be the predicted count obtained from the standard output of the evaluated method. For COBICount and other density map methods, $\widehat N_i$ is the sum of the predicted density map. Given $N$ evaluation images, MAE and RMSE are
\begin{align}
    \mathrm{MAE}
    &=
    \frac{1}{N}
    \sum_{i=1}^{N}
    \left|
    \widehat N_i-N_i^*
    \right|,
    \label{eq:exp_mae}\\
    \mathrm{RMSE}
    &=
    \sqrt{
    \frac{1}{N}
    \sum_{i=1}^{N}
    \left(
    \widehat N_i-N_i^*
    \right)^2
    }.
    \label{eq:exp_rmse}
\end{align}
MAE gives the average size of the count error. RMSE gives more weight to images with a large error. Lower values are better for both measures.

For the main experiment, mean target MAE (MT-MAE) summarizes the three unseen DOTA domains:
\begin{equation}
    \mathrm{MT\mbox{-}MAE}
    =
    \frac{
    \mathrm{MAE}_{LV}
    +
    \mathrm{MAE}_{SV}
    +
    \mathrm{MAE}_{Ship}
    }{3}.
    \label{eq:exp_mt_mae}
\end{equation}
The RSOC Building result is a source reference and is not included in this average. Later analyses add measures for the direction of count error and the locations of diagnostic points.

Methods that need a density map during training replace every source point with a normalized Gaussian response whose sum is one. This construction keeps the sum of the reference density map equal to the number of objects. Methods based on direct point prediction, probability, or distribution matching retain their own training objectives, and their counts are taken from their standard outputs. When a density map or point set is resized, its values or coordinates are adjusted so that the object count is preserved.

For every implemented method, the longer side of a target image is limited to 2000 pixels. This input rule is fixed before any target result is calculated, and all methods use the same preprocessing and input size within each target subset. No target label, count, box size, validation result, feature summary, or normalization statistic is used to choose that size. Each main result uses the same frozen checkpoint selected on the RSOC Building validation split; a separate checkpoint is not selected for each target.

The comparison includes MCNN~\cite{zhang2016mcnn}, CSRNet~\cite{li2018csrnet}, CAN~\cite{liu2019can}, Bayesian Loss~\cite{ma2019bayesian}, DM-Count~\cite{wang2020dmcount}, P2PNet~\cite{song2021p2pnet}, TransCrowd~\cite{liang2022transcrowd}, MobileNetV2 Counter~\cite{sandler2018mobilenetv2}, ResNet50 FPN Counter~\cite{he2016resnet,lin2017fpn}, MLDG-Count~\cite{li2018mldg}, MPCount~\cite{peng2024mpcount}, and BDRNet~\cite{guo2024bdrnet}. Together, they cover density map regression, direct point prediction, transformer counting, compact counters, counters with a feature pyramid, generic domain generalization, and density regression for remote sensing. MPCount is evaluated as a generic baseline under source-only counting, whereas BDRNet is evaluated as a baseline designed for remote sensing counting. Both official architectures are trained under source-only counting with RSOC Building. Their entries are results obtained after retraining under this setting rather than numbers published for their original datasets. None of the compared methods receives target data for adaptation or checkpoint selection.

\subsection{Implementation Details}
\label{subsec:exp_implementation}

\begin{table*}[!t]
\centering
\caption{Main comparison when RSOC Building is the source domain and DOTA Large Vehicle, DOTA Small Vehicle, and DOTA Ship are three unseen target domains. All training for the counting task and checkpoint selection use only RSOC Building. No target data are used for adaptation or checkpoint selection. MT-MAE is the average MAE over the three target domains and excludes the RSOC Building source result.}
\label{tab:main_results_final2}
\scriptsize
\setlength{\tabcolsep}{2.0pt}
\renewcommand{\arraystretch}{1.08}
\resizebox{\textwidth}{!}{%
\begin{tabular}{llccccccccccccc}
\toprule
\multirow{2}{*}{Type}
& \multirow{2}{*}{Method}
& \multirow{2}{*}{Year}
& \multirow{2}{*}{Adapt.}
& \multirow{2}{*}{Params (M)}
& \multirow{2}{*}{GFLOPs}
& \multicolumn{2}{c}{RSOC Building}
& \multicolumn{2}{c}{DOTA LV}
& \multicolumn{2}{c}{DOTA SV}
& \multicolumn{2}{c}{DOTA Ship}
& \multirow{2}{*}{MT-MAE} \\
\cmidrule(lr){7-8}
\cmidrule(lr){9-10}
\cmidrule(lr){11-12}
\cmidrule(lr){13-14}
& & & & & & MAE & RMSE & MAE & RMSE & MAE & RMSE & MAE & RMSE & \\
\midrule

\multirow{5}{*}{\begin{tabular}[c]{@{}c@{}}Standard /\\larger\end{tabular}}
& ResNet50 FPN Counter~\cite{he2016resnet,lin2017fpn}
& 2017 & $\times$ & 27.15 & 66.86
& 10.023 & 14.794 & 292.845 & 668.910 & 605.817 & 1322.535 & 590.810 & 882.637 & 496.491 \\

& CSRNet~\cite{li2018csrnet}
& 2018 & $\times$ & \textit{16.26} & \textit{27.07}
& 7.180 & 10.530 & 225.165 & 504.559 & 491.337 & 1118.768 & 494.861 & 687.234 & 403.788 \\

& CAN~\cite{liu2019can}
& 2019 & $\times$ & \textit{18.18} & --
& 9.120 & 13.380 & 76.182 & 96.163 & 415.150 & 1244.760 & 239.832 & 372.497 & 243.722 \\

& Bayesian Loss~\cite{ma2019bayesian}
& 2019 & $\times$ & \textit{21.50} & \textit{26.99}
& 28.950 & 32.930 & 583.585 & 826.571 & 872.199 & 1695.127 & 925.775 & 1316.316 & 793.853 \\

& P2PNet~\cite{song2021p2pnet}
& 2021 & $\times$ & \textit{19.20} & \textit{23.29}
& 7.460 & 10.340 & 124.993 & 158.005 & 411.526 & 1236.752 & 211.785 & 329.987 & 249.435 \\

\midrule
\multirow{4}{*}{Compact}
& MCNN~\cite{zhang2016mcnn}
& 2016 & $\times$ & \textit{0.13} & \textit{1.38}
& 12.130 & 17.350 & 479.545 & 613.845 & 630.747 & 1269.330 & 507.681 & 679.338 & 539.324 \\

& MobileNetV2 Counter~\cite{sandler2018mobilenetv2}
& 2018 & $\times$ & 5.47 & 2.40
& 9.893 & 14.935 & 442.202 & 896.202 & 784.845 & 1485.697 & 1079.061 & 1705.149 & 768.703 \\

& DM-Count~\cite{wang2020dmcount}
& 2020 & $\times$ & 0.83 & 18.80
& 9.427 & 13.688 & 202.395 & 299.733 & 473.152 & 1228.686 & 245.784 & 358.591 & 307.110 \\

& TransCrowd~\cite{liang2022transcrowd}
& 2022 & $\times$ & 0.97 & 0.71
& 8.580 & 12.510 & 134.181 & 180.581 & 415.060 & 1229.251 & 208.301 & 317.187 & 252.514 \\

\midrule
\multirow{2}{*}{Generic DG}
& MLDG-Count~\cite{li2018mldg}
& -- & $\times$ & 24.81 & 71.79
& 11.121 & 15.113 & 38.262 & 59.958 & 557.179 & 1371.708 & 275.569 & 416.210 & 290.337 \\

& MPCount~\cite{peng2024mpcount}
& 2024 & $\times$ & 33.23 & 143.79
& 12.818 & 17.732 & 45.092 & 68.044 & 490.854 & 1321.268 & 280.206 & 424.194 & 272.050 \\

\midrule
Remote sensing
& BDRNet~\cite{guo2024bdrnet}
& 2024 & $\times$ & 17.09 & 91.64
& 8.043 & 12.588 & 39.477 & 57.487 & 455.877 & 1298.813 & 264.399 & 401.384 & 253.251 \\

\midrule
\textbf{Ours}
& \textbf{COBICount}
& -- & $\times$ & \textbf{5.07} & \textbf{17.41}
& \textbf{6.574} & \textbf{9.753}
& \textbf{29.874} & \textbf{36.681}
& \textbf{327.045} & \textbf{941.040}
& \textbf{165.478} & \textbf{266.511}
& \textbf{174.132} \\

\bottomrule
\end{tabular}%
}

\vspace{0.5mm}

\footnotesize{$\times$ means that no target adaptation is used. DG means domain generalization. Params (M) gives the number of parameters in millions, and GFLOPs gives billions of floating point operations. Within the Params (M) and GFLOPs columns, \textit{italic values} are taken from the corresponding published papers and retain the settings reported there. Upright values are measured from the implemented models in the stated environment; their GFLOPs use a $512\times512$ input. ``--'' means that a value is unavailable or does not apply.}
\end{table*}

The experiments use Python 3.12.11, PyTorch 2.8.0, Torchvision 0.23.0, CUDA 12.9, and cuDNN. They run on an NVIDIA GeForce RTX 5070 Ti Laptop GPU with 11.94 GB memory, 24 CPU cores, and 31.44 GB system memory. Parameter counts and floating point operations are measured with a $512\times512$ input.

COBICount is trained for 80 epochs with AdamW, an optimizer that updates parameters from running averages of past gradients and applies weight decay. The initial learning rate is $1.8\times10^{-4}$, the weight decay is $10^{-4}$, the batch size is 6, and the random seed is 3407. Automatic mixed precision uses both lower and full precision arithmetic to reduce memory use. The $\ell_2$ norm of the gradient is limited to 5.0. An exponential moving average (EMA) copy of the model is also maintained with a decay of 0.999. This copy combines its previous parameters with the current parameters at every update and is used for source validation and final evaluation.

The relative and logarithmic count losses reach their full weights gradually during the first eight epochs. Direct source supervision for Audit starts at epoch 10. Source training uses $512\times512$ crops. With probability 0.85, a crop is centered near a randomly chosen source point, with at most 96 pixels of horizontal and vertical displacement. Otherwise, a random crop is used. Horizontal and vertical flips are each applied with probability 0.5. With probability 0.25, brightness and contrast factors are sampled from $[0.85,1.15]$, and color saturation is sampled from $[0.90,1.10]$. Source validation images are resized to $512\times512$. All images are normalized with the ImageNet channel means and standard deviations.

MPCount and BDRNet are trained on the source for 80 epochs with $512\times512$ source crops, seed 3407, automatic mixed precision, and checkpoint selection from source validation. Each uses a physical batch size of one, and gradients are accumulated over six steps, giving an effective batch size of six. Both use AdamW with a learning rate of $10^{-3}$ and a weight decay of $10^{-4}$; later changes to the learning rate follow the schedule in each released implementation. Both begin from their standard ImageNet initialization. Here, a density scale is the fixed constant used by a released implementation to rescale density values during training. MPCount uses the official deterministic \texttt{DGModel\_final}, a density scale of 1000, and its density, classification map, and memory consistency losses. BDRNet uses the official architecture, a density scale of 100, density regression, and its auxiliary object region output. In the released BDRNet training code, the intersection over union term measures region overlap, but it uses a hard threshold and is converted to a scalar, so it supplies no gradient. We retain its effective differentiable objective: density mean squared error plus $0.01$ times the sum of binary cross entropy and Dice loss for the auxiliary object region output. Dice loss also measures overlap between the predicted and reference object regions. For MPCount and BDRNet, the program first attempts inference on the complete resized image and uses fixed nonoverlapping $512\times512$ tiles only if a GPU memory error occurs. The selected checkpoint is unchanged across all three target domains. The architecture, fixed coefficients and thresholds, loss weights, point extraction settings, and input sizes are fixed implementation settings. None is selected or adjusted using target images, target labels, target statistics, or target evaluation results. In the main experiment, only RSOC Building is used to develop these settings and select the checkpoint. When another dataset is designated as the source, only the validation split of that source selects its checkpoint.

\subsection{Main Results}
\label{subsec:main_source_only_results}

Table~\ref{tab:main_results_final2} compares methods trained under the same rule for access to target data. It is not a ranking of every published remote sensing counter, because many published results train and test a separate model on each category. Those numbers are not directly comparable unless the methods are retrained with one source and no target data. Methods that use target images, generated target labels, target statistics, or target adaptation are also excluded~\cite{gong2024crossearth,ma2024deglgan,liu2024sfodrs}. MLDG-Count and MPCount provide generic references under source-only counting~\cite{li2018mldg,du2023domain,mansilla2021domain,peng2024mpcount}. MPCount was originally designed for crowd counting across datasets, so its row reports retraining of the official deterministic architecture on RSOC Building rather than the result published for its original crowd datasets. BDRNet provides a recent reference designed for remote sensing and is retrained under source-only counting~\cite{guo2024bdrnet}. The domain general crowd counting method in~\cite{du2023domain} remains outside the table because no corresponding result was produced under the present remote sensing protocol.

The source and target columns answer different questions. RSOC Building measures how well a method fits the category used for training, whereas the three DOTA columns measure the transfer of the same checkpoint selected from source validation to unavailable categories and scenes. A low source error therefore does not guarantee a low target error. P2PNet, for example, obtains a source MAE of 7.460 but an MT-MAE of 249.435. BDRNet shows the same gap, with a source MAE of 8.043 and an MT-MAE of 253.251. MPCount obtains 12.818 and 272.050, respectively.

COBICount obtains the lowest MT-MAE among the compared methods, at 174.132. The closest baseline is CAN at 243.722, corresponding to a 28.553\% reduction in MT-MAE. P2PNet, TransCrowd, and BDRNet obtain 249.435, 252.514, and 253.251, respectively. The two added 2024 baselines, BDRNet and MPCount, obtain MT-MAE values of 253.251 and 272.050. Relative to them, COBICount reduces MT-MAE by 31.241\% and 35.993\%, respectively. COBICount also gives the lowest source MAE and the lowest MAE on each of the three unseen target domains.

The recent baselines provide two further observations. BDRNet is both smaller and more accurate than MPCount under this protocol: it uses 17.09 million parameters and 91.64 GFLOPs, compared with 33.23 million and 143.79 GFLOPs for MPCount, and it gives a lower MAE on the source and every target domain. This pattern is consistent with the value of a counting design for remote sensing, but it does not isolate architecture as the only cause because the methods retain different objectives and learning rate schedules. Both methods transfer comparatively well to DOTA Large Vehicle, with MAEs of 39.477 and 45.092, but their errors increase on DOTA Small Vehicle and Ship. In particular, their Small Vehicle RMSE values reach 1298.813 and 1321.268, confirming that dense small objects and a few images with large errors remain difficult.

COBICount also remains imperfect. Its MAE/RMSE is 327.045/941.040 on DOTA Small Vehicle and 165.478/266.511 on DOTA Ship. After resizing, small vehicles occupy fewer output cells than source buildings, whereas parking lines, roads, shadows, water boundaries, and harbor structures remain visually strong. MLDG-Count, BDRNet, and MPCount obtain low Large Vehicle MAEs of 38.262, 39.477, and 45.092 but do not retain the same accuracy across the other two targets. These results indicate that source variation or a recent architecture for remote sensing alone does not determine whether a response lies on an object rather than on a repeated background pattern. The later ablation and diagnostic point analyses examine this distinction for the methods for which the required spatial diagnostic outputs were produced.

\subsection{Tests with Different Source Domains}
\label{subsec:cross_source_generalization}

The main comparison uses one source choice. Table~\ref{tab:source_swap_final2} tests two additional choices to determine whether COBICount is tied to RSOC Building. Each row names the single annotated domain used for training and checkpoint selection. Each group of columns names the domain used for evaluation after that checkpoint is fixed. An entry in which the row and column name the same domain is a source reference. Every other entry evaluates an unseen domain. For DIOR Airplane, only box centers are used as points; box size and boundaries are not used.

\begin{table}[!t]
\centering
\caption{Evaluation with three choices of source domain. Each row gives the single domain used for training and checkpoint selection. Each column gives the domain used for evaluation. Entries outside the matching source column measure transfer to an unseen domain without target adaptation.}
\label{tab:source_swap_final2}
\scriptsize
\setlength{\tabcolsep}{2.4pt}
\renewcommand{\arraystretch}{1.08}
\resizebox{\columnwidth}{!}{%
\begin{tabular}{l c | cc | cc | cc}
\toprule
\multirow{2}{*}{Source domain}
& \multirow{2}{*}{Adapt.}
& \multicolumn{2}{c|}{RSOC Building}
& \multicolumn{2}{c|}{DIOR Airplane}
& \multicolumn{2}{c}{DOTA Ship} \\
\cmidrule(lr){3-4}
\cmidrule(lr){5-6}
\cmidrule(lr){7-8}
& & MAE & RMSE & MAE & RMSE & MAE & RMSE \\
\midrule
RSOC Building
& $\times$
& 6.574 & 9.753
& 15.206 & 21.957
& 165.478 & 266.511 \\

DIOR Airplane
& $\times$
& 17.389 & 19.496
& 0.885 & 1.579
& 220.378 & 317.497 \\

DOTA Ship
& $\times$
& 11.278 & 16.373
& 49.170 & 50.339
& 81.980 & 141.044 \\
\bottomrule
\end{tabular}%
}
\vspace{0.6mm}

\parbox{\columnwidth}{\centering
\scriptsize
$\times$ means that no target adaptation is used. DIOR Airplane uses only box centers as point annotations.
}
\end{table}

The result changes when the training source changes. DIOR Airplane to RSOC Building gives 17.389/19.496 MAE/RMSE, showing that a model trained outside RSOC can still provide useful building counts. However, the same DIOR Airplane model gives 220.378/317.497 on DOTA Ship. In the reverse direction, the DOTA Ship model gives 49.170/50.339 on DIOR Airplane, compared with the DIOR Airplane source reference of 0.885/1.579. Transfer is also asymmetric between RSOC Building and DOTA Ship: the two directions give 165.478/266.511 and 11.278/16.373, respectively.

During training, source annotations show where counting responses should occur. From the source images around those annotations, the model also learns the usual width and shape of a response, the number of nearby objects, and the backgrounds that often appear around them. A new category can differ in each of these aspects. The table therefore supports two conclusions: COBICount is not restricted to RSOC Building as its source, but its accuracy still depends on how closely the source patterns match those in the unseen domain.

\FloatBarrier
\section{Ablation Study and Analysis}
\label{sec:ablation_analysis}

\subsection{Ablation of Model Parts}
\label{subsec:component_ablation_final}

Table~\ref{tab:ablation_final} tests which parts of COBICount contribute to the reported behavior. Every altered model is trained from source data, and its reported checkpoint is selected by RSOC Building validation MAE. The DOTA Proxy column evaluates this checkpoint under a controlled crop size. It is called a proxy because it samples fixed crops rather than complete DOTA images. The set contains $512\times512$ crops centered near a randomly selected DOTA reference point. The random seed and the allowed displacement are fixed, a crop can contain additional objects, and the number of crops drawn from each retained image is fixed. Images without an object of the selected category are excluded.

This sampling rule favors regions that contain targets and omits empty images. Its MAE and RMSE are therefore not comparable with the measures on complete DOTA images in Table~\ref{tab:main_results_final2}. The proxy values are used only for the ablation analysis. They do not enter the loss and do not select the checkpoint reported in Table~\ref{tab:ablation_final}; that checkpoint is selected from RSOC Building validation data.

The variants follow the order of the counting path. ``Density only'' uses the three routed density branches and the effective valid mask, while bypassing the other CE cues, CA, BI, and Audit supervision. ``w/o BI'' sets the BI retention gate $\mathbf V$ to its neutral value. ``w/o BG'' removes the learned background response $\mathbf B_{bg}$. ``w/o Line/Broad/Grid'' removes the three structure risk terms $\mathbf R_{line}$, $\mathbf R_{broad}$, and $\mathbf R_{grid}$. ``w/o Support'' removes the pooled support cue $\mathbf E_h$ from CE. ``w/o Center'' removes the center factor $\mathbf A_c$ from CA while retaining its other factors. ``w/o Audit'' removes the four channel Audit output and its source supervision.

\begin{table*}[!t]
\centering
\caption{Ablation of COBICount on RSOC Building and DOTA Proxy. Each row removes one model part or one related group. The reported checkpoint is selected by RSOC Building validation MAE. DOTA Proxy is used only to analyze the selected variants and does not select a reported checkpoint.}
\label{tab:ablation_final}
\scriptsize
\setlength{\tabcolsep}{3.0pt}
\renewcommand{\arraystretch}{1.10}
\resizebox{\textwidth}{!}{%
\begin{tabular}{lccccc rrrr}
\toprule
\multirow{2}{*}{Variant}
& \multicolumn{2}{c}{Bias Isolation}
& \makecell[c]{Candidate\\Evidence}
& \makecell[c]{Candidate\\Acceptance}
& \multirow{2}{*}{Audit}
& \multicolumn{2}{c}{RSOC Building}
& \multicolumn{2}{c}{DOTA Proxy} \\
\cmidrule(lr){2-3}
\cmidrule(lr){4-4}
\cmidrule(lr){5-5}
\cmidrule(lr){7-8}
\cmidrule(lr){9-10}
& BG & Line/Broad/Grid
& Support
& Center
&
& MAE & RMSE & MAE & RMSE \\
\midrule

Density only
& -- & -- & -- & -- & --
& 9.334 & 14.342 & 25.841 & 54.013 \\

w/o BI
& -- & -- & $\checkmark$ & $\checkmark$ & $\checkmark$
& 610.964 & 625.417 & 655.634 & 665.124 \\

w/o BG
& -- & $\checkmark$ & $\checkmark$ & $\checkmark$ & $\checkmark$
& 261.307 & 268.222 & 283.996 & 292.246 \\

w/o Line/Broad/Grid
& $\checkmark$ & -- & $\checkmark$ & $\checkmark$ & $\checkmark$
& 447.158 & 457.968 & 484.697 & 492.678 \\

w/o Support
& $\checkmark$ & $\checkmark$ & -- & $\checkmark$ & $\checkmark$
& 10.657 & 15.300 & 27.835 & 54.651 \\

w/o Center
& $\checkmark$ & $\checkmark$ & $\checkmark$ & -- & $\checkmark$
& 7.043 & 10.961 & 18.865 & 54.687 \\

w/o Audit
& $\checkmark$ & $\checkmark$ & $\checkmark$ & $\checkmark$ & --
& 7.708 & 11.322 & 17.977 & 54.728 \\

Full
& $\checkmark$ & $\checkmark$ & $\checkmark$ & $\checkmark$ & $\checkmark$
& \textbf{6.574} & \textbf{9.753} & \textbf{16.937} & \textbf{53.063} \\

\bottomrule
\end{tabular}%
}
\vspace{0.5mm}

\footnotesize{BG is the learned background response. Line/Broad/Grid denotes the three structure risk terms in BI. Support is the pooled support cue $\mathbf E_h$. Center is the center gate $\mathbf A_c$. Audit is an auxiliary output trained with source annotations; it does not gate the final density map. ``--'' means that a part is disabled or does not apply.}
\end{table*}

\begin{figure}[!t]
    \centering
    \includegraphics[width=\columnwidth]{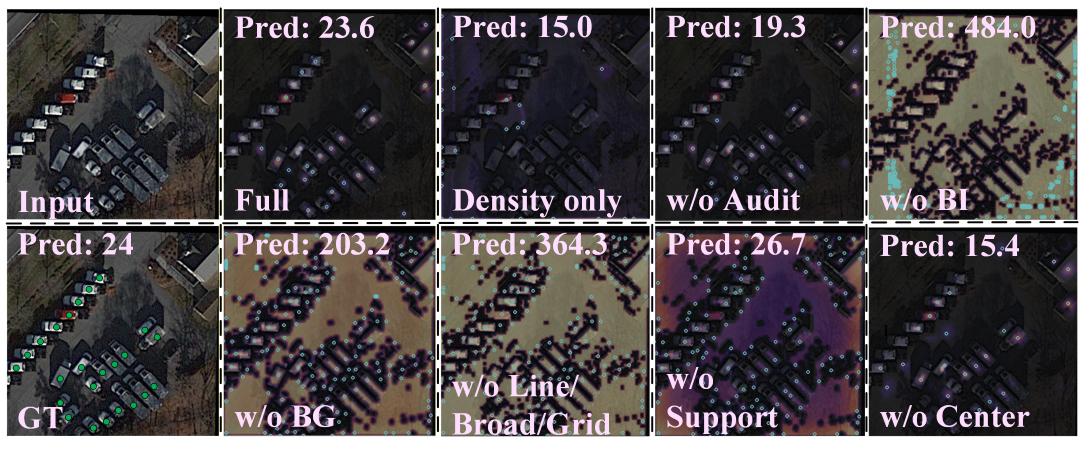}
    \caption{Examples from the ablation study when RSOC Building is the source and DOTA is unseen. The panels compare the input, reference points, the full model, and selected variants. They show how removing response generation, local acceptance, or structure suppression changes the final density map. Strong values on roads, parking structures, shadows, and repeated textures are responses without object support.}
    \label{fig:ablation_visualization}
\end{figure}

The variants reveal two different effects. First, ``Density only'' and ``w/o BI'' remove different sets of parts. ``Density only'' removes the three bounded CE cues together with CA and BI, leaving only the routed density path and the effective valid mask. By contrast, ``w/o BI'' retains the broader CE map and CA but fixes the retention gate $\mathbf V$ at its neutral value. Counting mass that BI would normally suppress and rescale can therefore accumulate. This difference explains why ``Density only'' remains numerically stable, whereas ``w/o BI'' raises RSOC Building MAE to 610.964 and DOTA Proxy MAE to 655.634. The result shows that BI controls both structure related responses and the scale of the summed density in the complete model; it does not show that each BI map identifies a named background category.

Second, the BI terms are not interchangeable. Removing BG raises DOTA Proxy MAE to 283.996, while removing the line, broad, and grid terms raises it to 484.697. Figure~\ref{fig:ablation_visualization} shows the corresponding increases on parking lines, road boundaries, shadows, and repeated structures. The table and figure therefore indicate that both the learned background response and the three structure risks are needed in the current formulation.

The CE support cue and the CA center gate have smaller but distinct effects. Removing $\mathbf E_h$ raises DOTA Proxy MAE from 16.937 to 27.835, which is consistent with a loss of weak responses that need evidence from a wider local area. Removing $\mathbf A_c$ raises RSOC Building MAE from 6.574 to 7.043 and DOTA Proxy MAE from 16.937 to 18.865. Removing Audit gives a more moderate change, as expected for an auxiliary output that does not gate the density map. The full model gives the lowest MAE on both sets.

\subsection{Prediction Bias}
\label{subsec:bias_analysis_final}

MAE and RMSE measure the size of an error, but not whether a model usually predicts too many or too few objects. To show this direction, we use Pred/GT, the ratio between the mean predicted count and the mean reference count:
\begin{equation}
    \mathrm{Pred/GT}
    =
    \frac{\frac{1}{N}\sum_{i=1}^{N}\widehat N_i}
    {\frac{1}{N}\sum_{i=1}^{N}N_i^*}
    =
    \frac{\sum_i\widehat N_i}{\sum_i N_i^*}.
    \label{eq:pred_gt_ratio}
\end{equation}
This is a ratio of two dataset means, not the mean of the ratios for individual images. It remains defined when some images contain no target, provided that the complete evaluation set contains at least one target. A value above one means that the total prediction is too high on average. A value below one means that it is too low.

\begin{table}[!t]
\centering
\caption{Pred/GT for selected methods when RSOC Building is the source and the three DOTA categories are unseen targets.}
\label{tab:bias_compact_final}
\scriptsize
\setlength{\tabcolsep}{3.0pt}
\renewcommand{\arraystretch}{1.08}
\resizebox{\columnwidth}{!}{%
\begin{tabular}{lcccc}
\toprule
Method & RSOC & DOTA LV & DOTA SV & DOTA Ship \\
\midrule
Bayesian Loss~\cite{ma2019bayesian} & 0.89 & 10.30 & 1.81 & 4.02 \\
P2PNet~\cite{song2021p2pnet} & 0.96 & 2.96 & 0.44 & 0.88 \\
CSRNet~\cite{li2018csrnet} & 0.96 & 4.50 & 1.21 & 2.29 \\
CAN~\cite{liu2019can} & 0.90 & 2.06 & 0.38 & 0.70 \\
DM-Count~\cite{wang2020dmcount} & 0.95 & 4.21 & 0.80 & 1.21 \\
TransCrowd~\cite{liang2022transcrowd} & 0.95 & 3.05 & 0.46 & 0.93 \\
MobileNetV2~\cite{sandler2018mobilenetv2} & 0.80 & 8.04 & 2.07 & 4.33 \\
MLDG-Count~\cite{li2018mldg} & 0.89 & 0.45 & 0.05 & 0.09 \\
COBICount & 0.97 & 1.78 & 0.52 & 0.92 \\
\bottomrule
\end{tabular}%
}
\vspace{0.5mm}

\footnotesize{Values above one indicate a prediction that is too high on average. Values below one indicate a prediction that is too low.}
\end{table}

The errors have a clear direction. Bayesian Loss, CSRNet, DM-Count, and MobileNetV2 produce much too large a total on DOTA Large Vehicle or DOTA Ship. Together with the visual maps, this pattern is consistent with count being assigned to roads, roof edges, water boundaries, harbor structures, or other strong background patterns. CAN, TransCrowd, P2PNet, and MLDG-Count instead produce too small a total on DOTA Small Vehicle, which is consistent with weak vehicle responses being missed after the large change in response size.

COBICount reduces these extremes but does not remove them. Its Pred/GT is 1.78 on DOTA Large Vehicle, so the total remains too high. On DOTA Ship, 0.92 is close to one compared with the large values of Bayesian Loss, CSRNet, and MobileNetV2. On DOTA Small Vehicle, 0.52 shows that COBICount still misses count, although the shortage is less severe than for CAN, TransCrowd, P2PNet, and MLDG-Count. In combination with the ablation results, these trends are consistent with CE retaining weak responses and BI limiting large amounts of background response. Pred/GT is therefore useful beside MAE because two methods with a similar absolute error can err in opposite directions.

\subsection{Locations of Diagnostic Points}
\label{subsec:candidate_diagnosis_final}

\begin{table*}[!t]
\centering
\caption{Location analysis for selected combinations of methods and domains using a matching radius of 32 pixels. All compared outputs are density maps. Recall@32 measures the coverage of reference objects, Precision@32 measures the fraction of diagnostic points supported by reference objects, and Diag/GT compares the numbers of diagnostic and reference points.}
\label{tab:candidate_quality_final}
\small
\setlength{\tabcolsep}{5.5pt}
\renewcommand{\arraystretch}{1.08}
\begin{tabular}{llrrr}
\toprule
Method & Evaluation domain & Recall@32 & Precision@32 & Diag/GT \\
\midrule
CAN~\cite{liu2019can}
& DOTA LV
& 0.279 & 0.079 & 3.408 \\

CAN~\cite{liu2019can}
& DOTA SV
& 0.193 & 0.186 & 0.714 \\

DM-Count~\cite{wang2020dmcount}
& DOTA SV
& 0.036 & 0.068 & 0.323 \\

MLDG-Count~\cite{li2018mldg}
& DOTA SV
& 0.060 & 0.071 & 1.623 \\

\midrule
COBICount
& RSOC Building
& 0.809 & 0.680 & 1.085 \\

COBICount
& DIOR Airplane
& 0.358 & 0.214 & 1.879 \\

COBICount
& DOTA LV
& 0.742 & 0.542 & 1.733 \\

COBICount
& DOTA SV
& 0.238 & 0.301 & 0.703 \\

COBICount
& DOTA Ship
& 0.614 & 0.394 & 1.273 \\
\bottomrule
\end{tabular}
\end{table*}

\begin{figure*}[!t]
    \centering
    \includegraphics[width=0.98\textwidth]{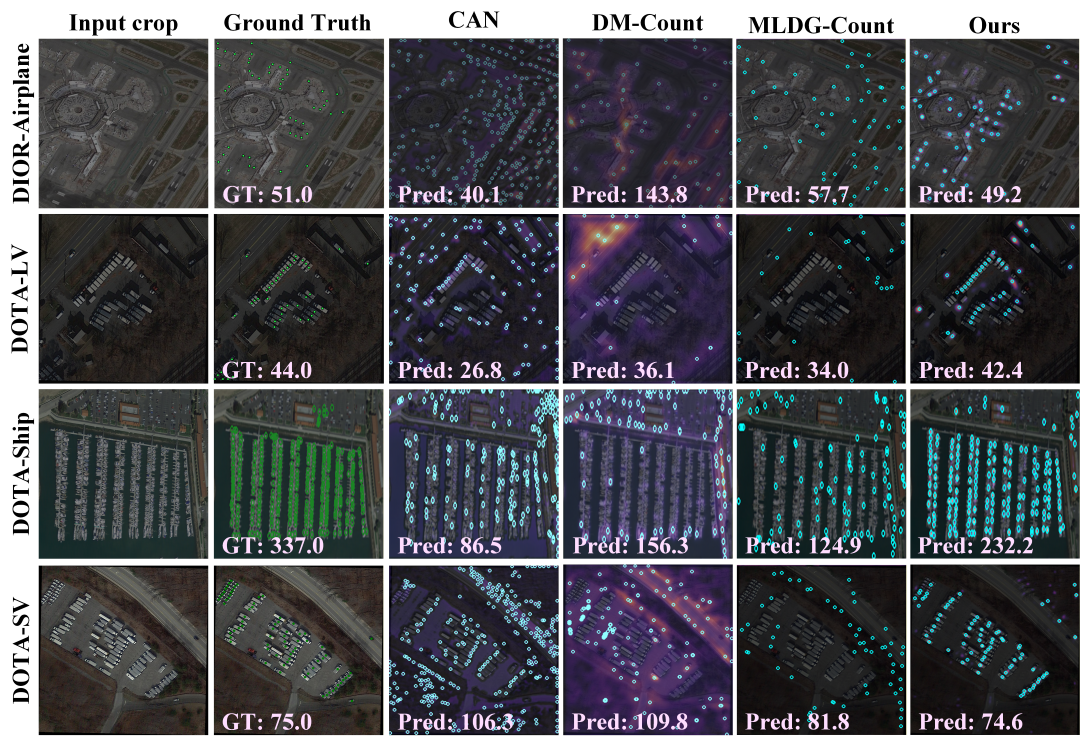}
    \caption{Response location examples on unseen DOTA and DIOR images. Each row contains an input crop, the reference points, and results from CAN, DM-Count, MLDG-Count, and COBICount. Green marks are reference points, and cyan marks are diagnostic points. The cyan marks are used only for visualization and location analysis; they are not native point predictions and do not produce the numerical count. As explained in Section~\ref{subsec:candidate_diagnosis_final}, COBICount and the baselines use fixed procedures suited to their respective output grids rather than one identical extraction rule. The displayed ``Pred'' value is the sum of the density map for the shown crop and is independent of the number of cyan marks.}
    \label{fig:candidate_transfer_visualization}
\end{figure*}

A correct total does not guarantee that counting responses lie on objects because missed objects and responses on background structures can partly cancel. We therefore use the diagnostic points defined in Section~\ref{subsec:diagnostic_interface}. They are extracted after the numerical count and are used only to compute Recall@32, Precision@32, and Diag/GT and to support the visualization in Fig.~\ref{fig:candidate_transfer_visualization}. Their number is not the predicted count; for every density map method in this comparison, the count remains the sum of the complete density map.

The location comparison uses a representative diagnostic subset consisting of CAN, DM-Count, MLDG-Count, and COBICount. These methods cover a density counter that uses context, a counter based on distribution matching, a generic reference under source-only counting, and the proposed model. MPCount and BDRNet are included in the numerical comparison in Table~\ref{tab:main_results_final2}, but no validated diagnostic point results are reported for them. P2PNet is not included because its standard output is a set of predicted object points rather than a density map. TransCrowd is not included because the implementation used in this study returns a numerical count but does not expose a compatible spatial map. This analysis is therefore not an exhaustive ranking of every method in Table~\ref{tab:main_results_final2}; it inspects response locations for the stated subset.

The numerical count results in Table~\ref{tab:main_results_final2} follow the input and preprocessing settings stated in Section~\ref{subsec:exp_metrics_comparison}; each count is taken from the standard numerical output of the evaluated method. The location analysis in this subsection is a separate diagnostic run. Here, large images are processed with nonoverlapping $512\times512$ tiles, and a border tile is padded when needed. All extracted points are then returned to the coordinates of the evaluated image. This tiling rule is used to obtain the diagnostic points in Table~\ref{tab:candidate_quality_final} and does not replace the count results in Table~\ref{tab:main_results_final2}. Figure~\ref{fig:candidate_transfer_visualization} shows selected crops, so its displayed ``Pred'' value refers only to the shown crop.

COBICount uses the fixed procedure defined in Section~\ref{subsec:diagnostic_interface} because this procedure was defined with the model and operates on the network output grid, whose stride is $4$. CAN, DM-Count, and MLDG-Count do not provide such a procedure, so one common rule is applied to their density maps at input image resolution. For these three methods, the threshold is the larger of $10^{-8}$ and $0.28$ times the largest map value. Nearby maxima within 6 pixels are suppressed, and at most 4096 points are retained. Neither procedure is changed for a target domain. Because the output grids and extraction rules differ, Table~\ref{tab:candidate_quality_final} is an analysis of the resulting response locations, not a controlled ranking under one identical point extraction rule.

For image $i$, let $\mathcal G_i$ contain the reference points and let $\mathcal Q_i$ contain the extracted diagnostic points. A reference point is covered when at least one diagnostic point lies within 32 pixels. A diagnostic point is supported when at least one reference point lies within the same radius. The reported measures pool the matched and total points over the full evaluation set:
\begin{equation}
\begin{aligned}
\mathrm{Recall}@32
&=
\frac{
\sum_i\sum_{\mathbf g\in\mathcal G_i}
\mathbf 1\!\left[
\min_{\mathbf q\in\mathcal Q_i}
\lVert\mathbf g-\mathbf q\rVert_2\le32
\right]
}{
\sum_i|\mathcal G_i|
},\\
\mathrm{Precision}@32
&=
\frac{
\sum_i\sum_{\mathbf q\in\mathcal Q_i}
\mathbf 1\!\left[
\min_{\mathbf g\in\mathcal G_i}
\lVert\mathbf q-\mathbf g\rVert_2\le32
\right]
}{
\sum_i|\mathcal Q_i|
},\\
\mathrm{Diag/GT}
&=
\frac{\sum_i|\mathcal Q_i|}{\sum_i|\mathcal G_i|}.
\end{aligned}
\label{eq:candidate_metrics_analysis}
\end{equation}
The minimum distance to an empty set is treated as infinity. Recall@32 is the fraction of reference objects covered by at least one diagnostic point. Precision@32 is the fraction of diagnostic points supported by at least one reference object. Diag/GT compares the total number of diagnostic points with the total number of reference points; Diag denotes the extracted diagnostic points, and GT denotes the reference points. The first two measures use independent nearest point tests, not a one to one assignment. Thus, two diagnostic points can be supported by the same object, and one diagnostic point can cover two nearby objects. If no diagnostic point is extracted, the denominator of Precision@32 is zero and the measure is undefined. We report this case as ``--'', while Recall@32 and Diag/GT are both zero.

Table~\ref{tab:candidate_quality_final} confirms that count accuracy and response location describe different properties. On DOTA Large Vehicle, CAN extracts 3.408 diagnostic points per reference point, but its Recall@32 and Precision@32 are only 0.279 and 0.079. Thus, many diagnostic points are away from annotated vehicles, while most reference vehicles remain uncovered. On DOTA Small Vehicle, its Diag/GT falls to 0.714, and its Recall@32 and Precision@32 are 0.193 and 0.186. CAN therefore extracts fewer points relative to the number of objects and still misses most small vehicles.

DM-Count extracts even fewer diagnostic points on DOTA Small Vehicle, with a Diag/GT of 0.323. Its Recall@32 is 0.036 and its Precision@32 is 0.068, showing that the small set of extracted points still has weak agreement with the reference points. MLDG-Count shows a different error. Its Diag/GT reaches 1.623, but its Recall@32 and Precision@32 are only 0.060 and 0.071. These results show that neither a large nor a small number of diagnostic points alone indicates reliable response locations.

On the RSOC Building source domain, COBICount obtains a Recall@32 of 0.809 and a Precision@32 of 0.680. After transfer, it retains values of 0.742 and 0.542 on DOTA Large Vehicle and 0.614 and 0.394 on DOTA Ship. DOTA Small Vehicle remains more difficult. COBICount and CAN extract similar numbers of diagnostic points relative to the reference points, with Diag/GT values of 0.703 and 0.714, respectively. However, COBICount gives higher Recall@32 and Precision@32, at 0.238 and 0.301 compared with 0.193 and 0.186 for CAN. On DIOR Airplane, its Diag/GT is 1.879 but its Precision@32 is 0.214, indicating that many diagnostic points remain away from airplane annotations.

Figure~\ref{fig:candidate_transfer_visualization} provides examples of these location errors. Some density maps produce many diagnostic points on parking lines, roads, water boundaries, harbor structures, or empty regions, whereas other maps leave many reference objects uncovered. Count error, Pred/GT, and the diagnostic point measures therefore answer three separate questions: how large the numerical error is, whether the total is too high or too low, and whether the local responses occur near objects.

\subsection{Boundary Analysis on DOTA Small Vehicle}
\label{subsec:failure_boundary_final}

DOTA Small Vehicle marks the main boundary of the present method. After the input is resized and passed to the network output grid with stride 4, a small vehicle may occupy only a few grid cells. A source building usually produces a wider response. At the same time, parking lines, road boundaries, lane marks, shadows, and enclosed empty spaces can remain clear over many cells. The local response from a true vehicle can therefore be weaker than the response from a repeated background pattern.

\begin{figure}[!h]
    \centering
    \includegraphics[width=\columnwidth]{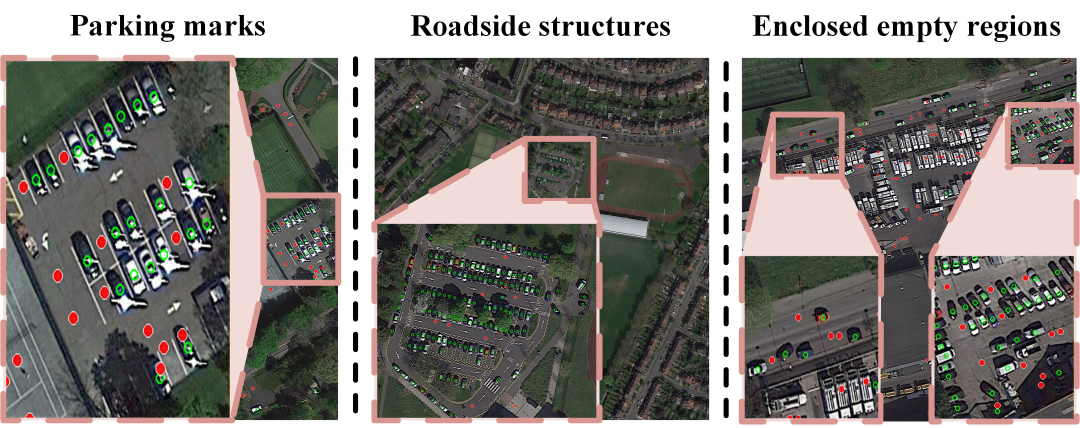}
    \caption{Representative errors on DOTA Small Vehicle. Manual inspection compares the input image, reference points, final density map, and diagnostic points. Strong responses occur on parking marks, roadside structures, and enclosed empty regions. These descriptions are made by the evaluator; they are not semantic classes predicted by Audit.}
    \label{fig:dota_sv_failure_boundary_final}
\end{figure}

This difference is an observed property of the evaluated images, not extra information supplied to the model. Target boxes are not used to set a response width, train the model, choose the checkpoint, or select an input size. Figure~\ref{fig:dota_sv_failure_boundary_final} groups the observed errors into responses on parking marks, roadside structures, and enclosed empty regions.

Physical object size alone does not explain these errors. What matters to the network is how much visible evidence remains on its network output grid. A small vehicle can lose most of its local pattern after resizing, whereas a long parking line or road edge remains strong. This problem is less severe for DOTA Large Vehicle because a large vehicle occupies more cells and usually forms a clearer local response.

CE, CA, and BI reduce some of these errors, but none removes the boundary completely. CE can produce little evidence for a very weak vehicle. CA can also retain a compact response from an enclosed empty region, and BI may not fully separate a weak vehicle from a clear parking or roadside pattern. Future work should generate stronger responses for objects that occupy few cells and provide stronger suppression for repeated source background patterns that resemble such objects.

\FloatBarrier
\section{Conclusion}
\label{sec:conclusion}

Many remote sensing counters are developed with annotated data from the category used at deployment, while methods for transfer often require target images or several annotated source domains. These requirements are difficult to meet when a new region, sensor, or object category must be processed without prior data collection. We therefore studied \emph{source-only counting}. This setting reduces the data needed before deployment and avoids further training for every new target. The difficulty is that a source model may respond not only to real objects but also to roads, parking patterns, roof boundaries, water boundaries, and other repeated structures. COBICount addresses this difficulty by separating response generation, acceptance, and background suppression. CE first produces a broad map of possible responses. CA then retains compact responses that agree with the patterns learned around source point annotations. BI finally reduces responses associated with learned background structures. Their outputs form the final density map, whose sum gives the predicted count. All supervision is obtained from source images and source point annotations. No target image, label, automatically generated label, feature summary, or model selection signal is used before the model is fixed. A separate Audit output is used only to inspect response patterns; it neither changes the final density map nor predicts named target categories. When trained on RSOC Building and applied directly to DOTA Large Vehicle, Small Vehicle, and Ship, COBICount obtains the lowest MT-MAE among the 12 compared baselines, including the 2024 MPCount and BDRNet methods, at 174.132. Relative to the closest baseline, CAN at 243.722, this is a 28.553\% reduction. The model contains 5.07 million parameters and requires 17.41 billion floating point operations for a $512\times512$ input. Tests that use DIOR Airplane or DOTA Ship as the source show that the model design can be trained from sources other than RSOC Building, although its accuracy still depends on the source data. The ablation results further show that CE, CA, and BI each contribute to the reported result. The experiments also show why count error alone provides incomplete evidence. MAE and RMSE measure the size of the numerical error, while Pred/GT shows whether the predicted total is generally too high or too low. Diagnostic point analysis examines whether local responses occur near reference points without changing the numerical count. On DOTA Large Vehicle, COBICount obtains a Recall@32 of 0.742 and a Precision@32 of 0.542. The lower values of 0.238 and 0.301 on DOTA Small Vehicle reveal the remaining difficulty in locating responses on dense small objects. These measures therefore separate numerical count accuracy from response location. DOTA Small Vehicle remains the main limitation of COBICount. After resizing, a small vehicle may occupy only a few output cells, while parking lines, road boundaries, shadows, and empty parking spaces can retain strong patterns. The model may therefore miss weak vehicle responses or retain responses from the background. Future work should strengthen responses for objects that occupy few cells, suppress repeated background structures without removing weak objects, and reduce computation for deployment on edge hardware. It should also examine how source data can be selected or combined without relying on target data.

\section*{Data and Code Availability}
The RSOC dataset is publicly available at \url{https://opendatalab.org.cn/OpenDataLab/RSOC}, and the DOTA dataset is publicly available at \url{https://captain-whu.github.io/DOTA/dataset.html}.\\
The code is available at \url{https://github.com/yixuxi22/COBICount}.

\bibliographystyle{IEEEtran}
\bibliography{reference}

@inproceedings{zhang2016mcnn,
  author    = {Yingying Zhang and Desen Zhou and Siqin Chen and Shenghua Gao and Yi Ma},
  title     = {Single-Image Crowd Counting via Multi-Column Convolutional Neural Network},
  booktitle = {Proceedings of the IEEE Conference on Computer Vision and Pattern Recognition},
  pages     = {589--597},
  year      = {2016}
}

@inproceedings{li2018csrnet,
  author    = {Yuhong Li and Xiaofan Zhang and Deming Chen},
  title     = {{CSRNet}: Dilated Convolutional Neural Networks for Understanding the Highly Congested Scenes},
  booktitle = {Proceedings of the IEEE Conference on Computer Vision and Pattern Recognition},
  pages     = {1091--1100},
  year      = {2018}
}

@inproceedings{liu2019can,
  author    = {Weizhe Liu and Mathieu Salzmann and Pascal Fua},
  title     = {Context-Aware Crowd Counting},
  booktitle = {Proceedings of the IEEE/CVF Conference on Computer Vision and Pattern Recognition},
  pages     = {5099--5108},
  year      = {2019}
}

@inproceedings{ma2019bayesian,
  author    = {Zhiheng Ma and Xing Wei and Xiaopeng Hong and Yihong Gong},
  title     = {Bayesian Loss for Crowd Count Estimation with Point Supervision},
  booktitle = {Proceedings of the IEEE/CVF International Conference on Computer Vision},
  pages     = {6142--6151},
  year      = {2019}
}

@inproceedings{wang2020dmcount,
  author    = {Wang, Boyu and Liu, Huidong and Samaras, Dimitris and Nguyen, Minh Hoai},
  title     = {Distribution Matching for Crowd Counting},
  booktitle = {Advances in Neural Information Processing Systems},
  volume    = {33},
  pages     = {1595--1607},
  year      = {2020}
}

@inproceedings{song2021p2pnet,
  author    = {Qingyu Song and Changan Wang and Zhengkai Jiang and Yabiao Wang and Ying Tai and Chengjie Wang and Jilin Li and Feiyue Huang and Yang Wu},
  title     = {Rethinking Counting and Localization in Crowds: A Purely Point-Based Framework},
  booktitle = {Proceedings of the IEEE/CVF International Conference on Computer Vision},
  pages     = {3365--3374},
  year      = {2021}
}

@article{liang2022transcrowd,
  author  = {Dingkang Liang and Xiwu Chen and Wei Xu and Yu Zhou and Xiang Bai},
  title   = {{TransCrowd}: Weakly-Supervised Crowd Counting with Transformers},
  journal = {Science China Information Sciences},
  volume  = {65},
  number  = {6},
  pages   = {160104},
  year    = {2022}
}

@article{gao2020counting,
  author  = {Guangshuai Gao and Qingjie Liu and Yunhong Wang},
  title   = {Counting From Sky: A Large-Scale Data Set for Remote Sensing Object Counting and a Benchmark Method},
  journal = {IEEE Transactions on Geoscience and Remote Sensing},
  volume  = {59},
  number  = {5},
  pages   = {3642--3655},
  year    = {2021},
  doi     = {10.1109/TGRS.2020.3020555}
}

@article{gao2022psgcnet,
  author  = {Guangshuai Gao and Qingjie Liu and Zhenghui Hu and Lu Li and Qi Wen and Yunhong Wang},
  title   = {{PSGCNet}: A Pyramidal Scale and Global Context Guided Network for Dense Object Counting in Remote-Sensing Images},
  journal = {IEEE Transactions on Geoscience and Remote Sensing},
  volume  = {60},
  pages   = {1--12},
  year    = {2022},
  doi     = {10.1109/TGRS.2022.3153946}
}

@article{guo2022tasnet,
  author  = {Xiangyu Guo and Marco Anisetti and Mingliang Gao and Gwanggil Jeon},
  title   = {Object Counting in Remote Sensing via Triple Attention and Scale-Aware Network},
  journal = {Remote Sensing},
  volume  = {14},
  number  = {24},
  pages   = {6363},
  year    = {2022},
  doi     = {10.3390/rs14246363}
}

@article{wang2025mscanet,
  author  = {Yixiao Wang and Zhenquan Wen and Xu Huang},
  title   = {{MSCA-Net}: Multiscale Chunked Attention Network for High-Resolution Satellite Stereo Matching},
  journal = {IEEE Journal of Selected Topics in Applied Earth Observations and Remote Sensing},
  volume  = {18},
  pages   = {27745--27763},
  year    = {2025},
  doi     = {10.1109/JSTARS.2025.3622164}
}

@inproceedings{xia2018dota,
  author    = {Gui-Song Xia and Xiang Bai and Jian Ding and Zhen Zhu and Serge Belongie and Jiebo Luo and Mihai Datcu and Marcello Pelillo and Liangpei Zhang},
  title     = {{DOTA}: A Large-Scale Dataset for Object Detection in Aerial Images},
  booktitle = {Proceedings of the IEEE Conference on Computer Vision and Pattern Recognition},
  pages     = {3974--3983},
  year      = {2018}
}

@article{lam2018xview,
  author  = {Darius Lam and Richard Kuzma and Kevin McGee and Samuel Dooley and Michael Laielli and Matthew Klaric and Yaroslav Bulatov and Brendan McCord},
  title   = {{xView}: Objects in Context in Overhead Imagery},
  journal = {arXiv preprint arXiv:1802.07856},
  year    = {2018}
}

@inproceedings{hsieh2017carpk,
  author    = {Meng-Ru Hsieh and Yen-Liang Lin and Winston H. Hsu},
  title     = {Drone-Based Object Counting by Spatially Regularized Regional Proposal Network},
  booktitle = {Proceedings of the IEEE International Conference on Computer Vision},
  pages     = {4145--4153},
  year      = {2017}
}

@inproceedings{sandler2018mobilenetv2,
  author    = {Mark Sandler and Andrew Howard and Menglong Zhu and Andrey Zhmoginov and Liang-Chieh Chen},
  title     = {{MobileNetV2}: Inverted Residuals and Linear Bottlenecks},
  booktitle = {Proceedings of the IEEE Conference on Computer Vision and Pattern Recognition},
  pages     = {4510--4520},
  year      = {2018}
}

@inproceedings{he2016resnet,
  author    = {Kaiming He and Xiangyu Zhang and Shaoqing Ren and Jian Sun},
  title     = {Deep Residual Learning for Image Recognition},
  booktitle = {Proceedings of the IEEE Conference on Computer Vision and Pattern Recognition},
  pages     = {770--778},
  year      = {2016}
}

@inproceedings{lin2017fpn,
  author    = {Tsung-Yi Lin and Piotr Doll{\'a}r and Ross Girshick and Kaiming He and Bharath Hariharan and Serge Belongie},
  title     = {Feature Pyramid Networks for Object Detection},
  booktitle = {Proceedings of the IEEE Conference on Computer Vision and Pattern Recognition},
  pages     = {2117--2125},
  year      = {2017}
}

@article{tuia2016domain,
  author  = {Devis Tuia and Claudio Persello and Lorenzo Bruzzone},
  title   = {Domain Adaptation for the Classification of Remote Sensing Data: An Overview of Recent Advances},
  journal = {IEEE Geoscience and Remote Sensing Magazine},
  volume  = {4},
  number  = {2},
  pages   = {41--57},
  year    = {2016},
  doi     = {10.1109/MGRS.2016.2548504}
}

@article{ganin2016dann,
  author  = {Yaroslav Ganin and Evgeniya Ustinova and Hana Ajakan and Pascal Germain and Hugo Larochelle and Fran{\c{c}}ois Laviolette and Mario Marchand and Victor Lempitsky},
  title   = {Domain-Adversarial Training of Neural Networks},
  journal = {Journal of Machine Learning Research},
  volume  = {17},
  number  = {59},
  pages   = {1--35},
  year    = {2016}
}

@article{du2023domain,
  author  = {Zhipeng Du and Jiankang Deng and Miaojing Shi},
  title   = {Domain-General Crowd Counting in Unseen Scenarios},
  journal = {Proceedings of the AAAI Conference on Artificial Intelligence},
  volume  = {37},
  number  = {1},
  pages   = {561--570},
  year    = {2023},
  doi     = {10.1609/aaai.v37i1.25131}
}

@inproceedings{peng2024mpcount,
  author    = {Zhuoxuan Peng and S.-H. Gary Chan},
  title     = {Single Domain Generalization for Crowd Counting},
  booktitle = {Proceedings of the IEEE/CVF Conference on Computer Vision and Pattern Recognition},
  pages     = {28025--28034},
  year      = {2024}
}

@inproceedings{radford2021clip,
  author    = {Alec Radford and Jong Wook Kim and Chris Hallacy and Aditya Ramesh and Gabriel Goh and Sandhini Agarwal and Girish Sastry and Amanda Askell and Pamela Mishkin and Jack Clark and Gretchen Krueger and Ilya Sutskever},
  title     = {Learning Transferable Visual Models From Natural Language Supervision},
  booktitle = {Proceedings of the International Conference on Machine Learning},
  pages     = {8748--8763},
  year      = {2021}
}

@inproceedings{kirillov2023sam,
  author    = {Alexander Kirillov and Eric Mintun and Nikhila Ravi and Hanzi Mao and Chloe Rolland and Laura Gustafson and Tete Xiao and Spencer Whitehead and Alexander C. Berg and Wan-Yen Lo and Piotr Doll{\'a}r and Ross Girshick},
  title     = {Segment Anything},
  booktitle = {Proceedings of the IEEE/CVF International Conference on Computer Vision},
  pages     = {4015--4026},
  year      = {2023}
}

@inproceedings{selvaraju2017gradcam,
  author    = {Ramprasaath R. Selvaraju and Michael Cogswell and Abhishek Das and Ramakrishna Vedantam and Devi Parikh and Dhruv Batra},
  title     = {Grad-{CAM}: Visual Explanations From Deep Networks via Gradient-Based Localization},
  booktitle = {Proceedings of the IEEE International Conference on Computer Vision},
  pages     = {618--626},
  year      = {2017}
}

@article{liu2024remoteclip,
  author  = {Fan Liu and Delong Chen and Zhangqingyun Guan and Xiaocong Zhou and Jiale Zhu and Qiaolin Ye and Liyong Fu and Jun Zhou},
  title   = {{RemoteCLIP}: A Vision Language Foundation Model for Remote Sensing},
  journal = {IEEE Transactions on Geoscience and Remote Sensing},
  volume  = {62},
  pages   = {1--16},
  year    = {2024},
  doi     = {10.1109/TGRS.2024.3390838}
}

@article{zhang2024georsclip,
  author  = {Zilun Zhang and Tiancheng Zhao and Yulong Guo and Jianwei Yin},
  title   = {{RS5M} and {GeoRSCLIP}: A Large-Scale Vision-Language Dataset and a Large Vision-Language Model for Remote Sensing},
  journal = {IEEE Transactions on Geoscience and Remote Sensing},
  volume  = {62},
  pages   = {1--23},
  year    = {2024},
  doi     = {10.1109/TGRS.2024.3449154}
}

@article{chen2023rsprompter,
  author  = {Keyan Chen and Chenyang Liu and Hao Chen and Haotian Zhang and Wenyuan Li and Zhengxia Zou and Zhenwei Shi},
  title   = {{RSPrompter}: Learning to Prompt for Remote Sensing Instance Segmentation Based on Visual Foundation Model},
  journal = {IEEE Transactions on Geoscience and Remote Sensing},
  volume  = {62},
  pages   = {1--17},
  year    = {2024},
  doi     = {10.1109/TGRS.2024.3356074}
}

@article{shen2025edgecount,
  author  = {Zhilong Shen and Guoquan Li and Ruiyang Xia and Hongying Meng and Zhengwen Huang},
  title   = {A Lightweight Object Counting Network Based on Density Map Knowledge Distillation},
  journal = {IEEE Transactions on Circuits and Systems for Video Technology},
  volume  = {35},
  number  = {2},
  pages   = {1492--1505},
  year    = {2025},
  doi     = {10.1109/TCSVT.2024.3469933}
}

@article{li2020dior,
author  = {Ke Li and Gang Wan and Gong Cheng and Liqiu Meng and Junwei Han},
title   = {Object Detection in Optical Remote Sensing Images: A Survey and A New Benchmark},
journal = {ISPRS Journal of Photogrammetry and Remote Sensing},
volume  = {159},
pages   = {296--307},
year    = {2020}
}

@article{sun2022fair1m,
author  = {Xian Sun and Peijin Wang and Zhiyuan Yan and Feng Xu and Ruiping Wang and Wenhui Diao and Jin Chen and Jihao Li and Yingchao Feng and Tao Xu and Martin Weinmann and Stefan Hinz and Cheng Wang and Kun Fu},
title   = {{FAIR1M}: A Benchmark Dataset for Fine-Grained Object Recognition in High-Resolution Remote Sensing Imagery},
journal = {ISPRS Journal of Photogrammetry and Remote Sensing},
volume  = {184},
pages   = {116--130},
year    = {2022}
}

@inproceedings{li2018mldg,
author    = {Da Li and Yongxin Yang and Yi-Zhe Song and Timothy M. Hospedales},
title     = {Learning to Generalize: Meta-Learning for Domain Generalization},
booktitle = {Proceedings of the AAAI Conference on Artificial Intelligence},
year      = {2018}
}

@inproceedings{cong2022satmae,
author  = {Yezhen Cong and Samar Khanna and Chenlin Meng and Patrick Liu and Erik Rozi and Yutong He and Marshall Burke and David B. Lobell and Stefano Ermon},
title   = {{SatMAE}: Pre-training Transformers for Temporal and Multi-Spectral Satellite Imagery},
booktitle = {Advances in Neural Information Processing Systems},
volume    = {35},
pages     = {197--211},
year      = {2022}
}

@inproceedings{reed2023scalemae,
author    = {Colorado J. Reed and Ritwik Gupta and Shufan Li and Sarah Brockman and Christopher Funk and Brian Clipp and Kurt Keutzer and Salvatore Candido and Matt Uyttendaele and Trevor Darrell},
title     = {Scale-{MAE}: A Scale-Aware Masked Autoencoder for Multiscale Geospatial Representation Learning},
booktitle = {Proceedings of the IEEE/CVF International Conference on Computer Vision},
pages     = {4088--4099},
year      = {2023}
}

@inproceedings{noman2024satmaepp,
  author    = {Mubashir Noman and Muzammal Naseer and Hisham Cholakkal and Rao Muhammad Anwer and Salman Khan and Fahad Shahbaz Khan},
  title     = {Rethinking Transformers Pre-training for Multi-Spectral Satellite Imagery},
  booktitle = {Proceedings of the IEEE/CVF Conference on Computer Vision and Pattern Recognition},
  pages     = {27811--27819},
  year      = {2024}
}

@article{gong2024crossearth,
author  = {Ziyang Gong and Zhixiang Wei and Di Wang and Xianzheng Ma and Hongruixuan Chen and Yuru Jia and Yupeng Deng and Zhenming Ji and Xiangwei Zhu and Naoto Yokoya and Jing Zhang and Bo Du and Liangpei Zhang},
title   = {{CrossEarth}: Geospatial Vision Foundation Model for Domain Generalizable Remote Sensing Semantic Segmentation},
journal = {arXiv preprint arXiv:2410.22629v1},
year    = {2024},
url     = {https://arxiv.org/abs/2410.22629v1}
}

@article{pan2024lae,
  author  = {Pan, Jiancheng and Liu, Yanxing and Fu, Yuqian and Ma, Muyuan and Li, Jiahao and Paudel, Danda Pani and Van Gool, Luc and Huang, Xiaomeng},
  title   = {Locate Anything on Earth: Advancing Open-Vocabulary Object Detection for Remote Sensing Community},
  journal = {Proceedings of the AAAI Conference on Artificial Intelligence},
  volume  = {39},
  number  = {6},
  pages   = {6281--6289},
  year    = {2025},
  doi     = {10.1609/aaai.v39i6.32672}
}

@inproceedings{wang2023samrs,
author  = {Di Wang and Jing Zhang and Bo Du and Minqiang Xu and Lin Liu and Dacheng Tao and Liangpei Zhang},
title   = {{SAMRS}: Scaling-up Remote Sensing Segmentation Dataset with Segment Anything Model},
booktitle = {Advances in Neural Information Processing Systems},
volume    = {36},
pages     = {8815--8827},
year      = {2023}
}

@article{osco2023samrsapp,
  author  = {Osco, Lucas Prado and Wu, Qiusheng and de Lemos, Eduardo Lopes and Gon{\c{c}}alves, Wesley Nunes and Ramos, Ana Paula Marques and Li, Jonathan and Marcato, Junior, Jos{\'e}},
  title   = {The Segment Anything Model ({SAM}) for Remote Sensing Applications: From Zero to One Shot},
  journal = {International Journal of Applied Earth Observation and Geoinformation},
  volume  = {124},
  pages   = {103540},
  year    = {2023},
  doi     = {10.1016/j.jag.2023.103540}
}

@article{geirhos2020shortcut,
  author  = {Robert Geirhos and J{\"o}rn-Henrik Jacobsen and Claudio Michaelis and Richard Zemel and Wieland Brendel and Matthias Bethge and Felix A. Wichmann},
  title   = {Shortcut Learning in Deep Neural Networks},
  journal = {Nature Machine Intelligence},
  volume  = {2},
  number  = {11},
  pages   = {665--673},
  year    = {2020}
}

@inproceedings{chen2025urm,
  author  = {Xianing Chen and Si Huo and Borui Jiang and Hailin Hu and Xinghao Chen},
  title   = {Single Domain Generalization for Few-Shot Counting via Universal Representation Matching},
  booktitle = {Proceedings of the IEEE/CVF Conference on Computer Vision and Pattern Recognition},
  pages     = {4639--4649},
  year      = {2025}
}

@article{wieland2024s1s2water,
  author  = {Marc Wieland and Florian Fichtner and Sandro Martinis and Sandro Groth and Christian Krullikowski and Simon Plank and Mahdi Motagh},
  title   = {{S1S2-Water}: A Global Dataset for Semantic Segmentation of Water Bodies From Sentinel-1 and Sentinel-2 Satellite Images},
  journal = {IEEE Journal of Selected Topics in Applied Earth Observations and Remote Sensing},
  volume  = {17},
  pages   = {1084--1099},
  year    = {2024},
  doi     = {10.1109/JSTARS.2023.3333969}
}

@inproceedings{mansilla2021domain,
  author    = {Mansilla, Lucas and Echeveste, Rodrigo and Milone, Diego H. and Ferrante, Enzo},
  title     = {Domain Generalization via Gradient Surgery},
  booktitle = {Proceedings of the IEEE/CVF International Conference on Computer Vision},
  pages     = {6630--6638},
  year      = {2021}
}

@article{ma2024deglgan,
  author  = {Xianping Ma and Xiaokang Zhang and Xingchen Ding and Man-On Pun and Siwei Ma},
  title   = {Decomposition-based Unsupervised Domain Adaptation for Remote Sensing Image Semantic Segmentation},
  journal = {IEEE Transactions on Geoscience and Remote Sensing},
  volume  = {62},
  pages   = {1--18},
  year    = {2024},
  doi     = {10.1109/TGRS.2024.3483283}
}

@article{liu2024sfodrs,
  author  = {Weixing Liu and Jun Liu and Xin Su and Han Nie and Bin Luo},
  title   = {Source-free Domain Adaptive Object Detection in Remote Sensing Images},
  journal = {arXiv preprint arXiv:2401.17916},
  year    = {2024}
}

@article{guo2024bdrnet,
  author  = {Guo, Haojie and Gao, Junyu and Yuan, Yuan},
  title   = {Balanced Density Regression Network for Remote Sensing Object Counting},
  journal = {IEEE Transactions on Geoscience and Remote Sensing},
  year    = {2024},
  volume  = {62},
  pages   = {1--13},
  doi     = {10.1109/TGRS.2024.3402271}
}

\end{document}